%% file: iclr2027_conference.tex
\documentclass{article} 
\usepackage{iclr2027_conference,times}

\input{math_commands.tex}

\usepackage{xcolor}
\usepackage{hyperref}
\usepackage{url}
\usepackage{algorithm}
\usepackage{algorithmic}
\usepackage{float}
\usepackage{booktabs}
\usepackage{multirow}
\usepackage{graphicx}
\usepackage{caption}
\usepackage{array}
\usepackage{amsthm}
\usepackage{enumitem}
\usepackage{tikz}
\usepackage[export]{adjustbox}
\usetikzlibrary{arrows.meta,backgrounds,calc,fit,positioning}
\theoremstyle{definition}
\newtheorem{definition}{Definition}
\hypersetup{
  citebordercolor={0 1 0},
  linkbordercolor={1 1 1},
  urlbordercolor={1 1 1}
}

\title{VLALight: A Vision-Language-Action Model for Traffic Signal Control}

\author{
Pan Zhang$^{1}$\thanks{Equal contribution.}
\quad
Siqi Lai$^{1}$\footnotemark[1]
\quad
Kemu Dong$^{2}$
\quad
Hao Liu$^{1}$\thanks{Corresponding author.}
\\[0.5em]
$^{1}$The Hong Kong University of Science and Technology (Guangzhou)\\
$^{2}$Dalian University of Technology\\[0.25em]
\small\texttt{pzhang521@connect.hkust-gz.edu.cn}
}

\newcommand{\siqi}[1]{{\color{black}{#1}}}

\iclrfinalcopy 
\begin{document}

\maketitle
\fancyhead{}
\renewcommand{\headrulewidth}{0pt}

\begin{abstract}
    Traffic signal control (TSC) is essential for improving urban mobility and reducing congestion. Although roadside cameras are widely deployed at signalized intersections and provide rich visual observations of evolving traffic, existing TSC methods typically rely on manually engineered traffic states or separate perception modules, creating a gap between physical observations and control decisions. We present \textbf{VLALight}, the first vision-language-action (VLA) model for end-to-end traffic signal control from multi-view roadside videos. VLALight directly maps visual observations to coordinated signal actions through multi-target spatiotemporal traffic reasoning and topology-aware cooperative perception across intersections. To establish this capability, we develop a two-stage supervised cold-start training strategy for visual traffic understanding and signal decision-making, followed by cooperative agentic reinforcement learning that jointly optimizes local control and network-wide traffic efficiency. Furthermore, VLALight introduces adaptive fast and slow reasoning modes, enabling the policy to allocate deeper reasoning only when additional deliberation provides sufficient control benefits. Through balanced mode-aware rollouts and relative advantage optimization, VLALight learns to trade off decision quality and inference cost. Extensive experiments on seven real-world traffic-flow datasets across three urban networks demonstrate that VLALight consistently outperforms transportation-based, RL-based, and LLM/VLM-based baselines. Ablation studies validate the effectiveness of cooperative perception, network-level optimization, and adaptive reasoning. These results demonstrate the potential of VLA models for real-world physical traffic control. Our project is available at \url{https://github.com/usail-hkust/VLALight.git}.
\end{abstract}

\vspace{-5pt}
\section{Introduction}

\siqi{Traffic signal control (TSC) is a fundamental component of urban traffic management. By regulating conflicting movements at intersections, it directly affects vehicle delay, network throughput, and traffic-related emissions. Meanwhile, roadside cameras have become increasingly prevalent at signalized intersections, providing continuous and information-rich observations of vehicle movements, lane occupancy, and surrounding road conditions~\citep{wang2021roadside}. Compared with conventional detector-based measurements (e.g., underground coils), such visual observations preserve substantially richer information about the physical traffic scene and are already available in many real-world deployments. This creates a natural opportunity to develop traffic signal controllers that directly perceive and act upon visual observations. However, despite the growing availability of roadside vision, existing TSC systems largely operate on manually engineered traffic states rather than raw physical observations, leaving end-to-end visual traffic control largely unexplored.}

\siqi{Existing traffic signal control research has evolved from transportation-engineering methods toward learning-based control. Classical approaches, including signal timing optimization~\citep{koonce2008timing}, SCOOT~\citep{hunt1982scoot}, and MaxPressure~\citep{varaiya2013maxpressure}, are efficient and interpretable but rely on handcrafted control logic and simplified traffic-flow assumptions~\citep{qadri2020review,weietal2021survey}. Reinforcement learning (RL) improves adaptability by learning control policies from environmental feedback~\citep{abdulhai2003rl,wei2018intellilight,zheng2019frap,wei2019presslight,chen2020mplight,wei2019colight,wu2021dynstgat,yu2020macar,ruan2024coslight}. More recently, LLM-based methods such as LLMLight~\citep{lai2025llmlight}, Traffic-R1~\citep{zou2026trafficr1}, and CoLLMLight~\citep{yuan2026collmlight} introduce stronger reasoning and interpretability. However, both RL- and LLM-based approaches generally rely on carefully constructed traffic states, such as queue lengths, vehicle counts, occupancy, and lane pressure. Although such states are readily available in traffic simulators, they can be incomplete or unreliable in real-world deployments because of limited sensor coverage, sensor failures, and communication interruptions~\citep{mei2023missing}. Vision-language models (VLMs) provide a promising way to bridge this perception--control gap by directly accessing traffic observations from traffic cameras, which preserve vehicle motion, road geometry, and evolving traffic patterns beyond predefined traffic measures~\citep{zhou2023vlmsurvey}. However, existing VLM-based TSC methods primarily focus on scene understanding or high-level assistance, while delegating final signal actions to separate RL or language-model controllers~\citep{wang2025vlmlight}. Consequently, end-to-end signal control directly from physical visual observations remains largely underexplored.}



\siqi{Vision-language-action (VLA) models have recently emerged as a unified paradigm for embodied decision-making by grounding visual-language representations into executable actions~\citep{brohan2023rt2,kim2024openvla}. This formulation is naturally suited for traffic signal control, as it enables an agent to directly interpret physical traffic scenes, reason about their dynamics, and generate signal actions within a unified policy. However, extending VLA to TSC introduces three substantial challenges. First, videos from traffic cameras contain dense and continuously evolving interactions among vehicles, lanes, and movement directions. Effective control therefore requires complex multi-target spatiotemporal reasoning to identify critical traffic dynamics and anticipate future evolution. Second, TSC is inherently a network-wide optimization problem: a signal decision at one intersection can reshape downstream arrivals, queue propagation, and spillback. A network-aware VLA controller needs to reason over spatially connected intersections and coordinate actions beyond local observations. Finally, the inference latency must satisfy strict efficiency requirements. While complex congestion may benefit from deeper deliberation, routine conditions should avoid unnecessary computation. The controller must consequently adapt its reasoning effort to traffic complexity while balancing decision quality and inference latency.}

\siqi{
To address these challenges, we propose \textit{VLALight}, an end-to-end vision-language-action model for traffic signal control that directly maps physical visual observations to executable signal actions. Given multi-view intersection videos, VLALight first performs multi-target traffic perception and spatiotemporal reasoning to capture evolving vehicle movements, queue dynamics, and traffic trends. To address the networked nature of TSC, it further incorporates observations from neighboring intersections through topology-aware cooperative perception, enabling coordinated signal decisions beyond isolated local control. To progressively establish these capabilities, we adopt a two-stage supervised fine-tuning strategy that separately develops spatiotemporal traffic understanding and executable signal decision-making. Building on this initialization, we introduce a cooperative agentic RL framework that optimizes network-wide control through environment interaction. The training objective jointly considers immediate local queue reduction and longer-horizon network-level traffic improvement, encouraging actions that improve both individual intersections and overall traffic efficiency. Furthermore, VLALight explicitly explores balanced fast- and slow-reasoning modes during RL rollouts, allowing the policy to compare direct responses with deeper deliberation under different traffic conditions. Through mode-wise advantage normalization, VLALight learns when additional reasoning provides sufficient control benefits to justify its computational overhead. As a result, VLALight unifies visual traffic understanding, cooperative multi-intersection control, and adaptive reasoning efficiency within a single VLA model.
}


\siqi{Our contributions are threefold. (1) We introduce VLALight, an end-to-end VLA model that directly connects physical roadside observations with spatiotemporal traffic reasoning and coordinated signal control. To our knowledge, it is the first VLA model in traffic signal control. (2) We develop a cooperative training framework with staged supervised fine-tuning, followed by agentic reinforcement learning for coordinated multi-intersection control and cost-effective adaptive reasoning. (3) Extensive experiments on seven real-world traffic datasets demonstrate that VLALight consistently improves network-wide traffic efficiency and cost-effectiveness across diverse traffic conditions.
}


\vspace{-5pt}
\section{Preliminaries}

\siqi{
\begin{definition}[Road Network]
We represent an urban road network as a directed graph $\mathcal{G}=(\mathcal{I},\mathcal{L})$, where $\mathcal{I}$ is the set of signalized intersections and $\mathcal{L}$ is the set of directed lane connections between them. Each intersection has incoming and outgoing lanes, and each lane permits one or more traffic movements, such as going straight, turning left, or turning right. The graph captures the spatial pathways along which vehicles, congestion, and the effects of signal decisions propagate.
\end{definition}

\begin{definition}[Vision-Based Traffic Signal Control]
Vision-based traffic signal control is formulated as a partially observable Markov decision process (POMDP)
$\langle\mathcal{S},\mathcal{O},\mathcal{A},R\rangle$:
\begin{itemize}[leftmargin=0.5cm,itemsep=0pt,topsep=0.2em,parsep=0pt]
    \item \textbf{State.} $\mathcal{S}$ is the latent traffic-state space. The local state $s_i^t\in\mathcal{S}$ at intersection $i$ may include queue lengths, approaching-vehicle counts, waiting times, vehicle speeds, traffic density, and the current signal phase. The controller does not directly observe this complete state.
    \item \textbf{Observation.} $\mathcal{O}$ is the visual-observation space. Let $x_{i,d}^{t,k}$ denote the $k$-th image frame captured during control interval $t$ by the roadside camera facing approach $d\in\{E,W,N,S\}$. The observation available to intersection $i$ is the collection of frame sequences
    $o_i^t=\{x_{i,d}^{t,k}\}_{d\in\{E,W,N,S\},\,k=1,\dots,K}$.
    \item \textbf{Action.} The feasible action set at intersection $i$ is $\mathcal{A}_i=\{a_{i,1},\ldots,a_{i,m_i}\}$. An action $a_i^t\in\mathcal{A}_i$ selects a signal phase, i.e., a non-conflicting set of traffic movements that receives a green signal during the next control interval. The joint action space is $\mathcal{A}=\prod_{i\in\mathcal{I}}\mathcal{A}_i$.
    \item \textbf{Reward.} $R$ evaluates the traffic outcome after an action. The network reward $R^t=R(s^t,\mathbf{a}^t,s^{t+1})$ measures system-wide traffic efficiency (e.g., queue length, travel time).
\end{itemize}
\end{definition}

We consider a shared vision-based policy $\pi_\theta$ across intersections. At each signal-switching time step $t$, intersection $i$ selects an action from its local frame observation $o_i^t$, neighbouring intersection context $c_i^t$, and its feasible phase set $\mathcal{A}_i$. The policy is optimized directly from visual observations to maximize long-term network-wide traffic efficiency:
\begin{equation}
    \mathbf{a}^t=\{\pi_\theta(o_i^t,c_i^t,\mathcal{A}_i)\}_{i\in\mathcal{I}}, \qquad
    \theta^*=\arg\max_\theta\,
    \mathbb{E}_{\pi_\theta}\!\left[\sum_{t=0}^{T-1}R^t(s^t,\mathbf{a}^t,s^{t+1})\right].
\end{equation}
}

\begin{figure}[t]
    \centering
    \includegraphics[page=3,width=\linewidth]{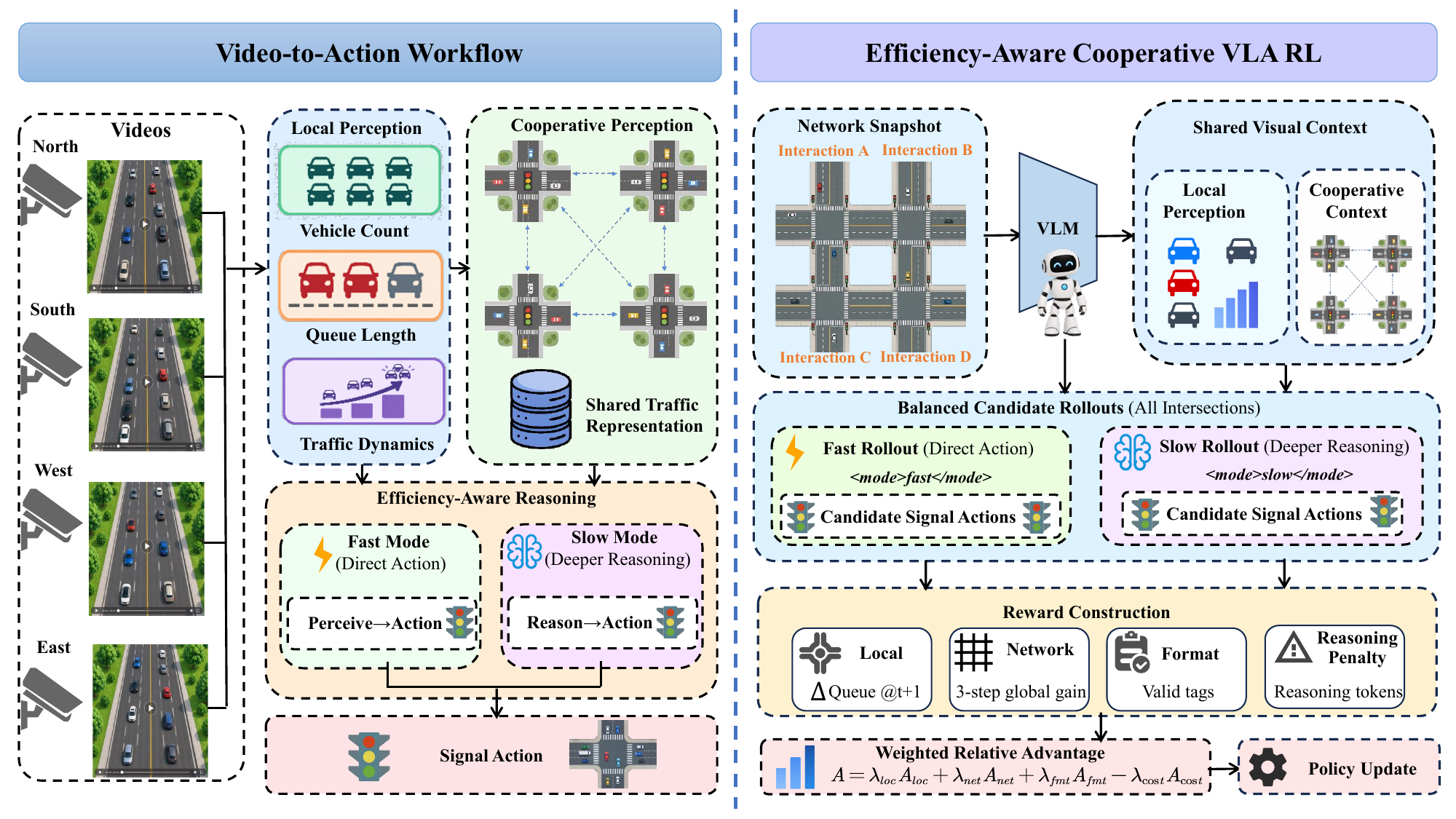}
    \caption{The overview of VLALight.}
    \label{fig:vlalight_overview}
\vspace{-10pt}
\end{figure}

\vspace{-5pt}
\siqi{
\section{Method}

We present VLALight, an end-to-end VLA model for cooperative traffic signal control. As illustrated in Figure~\ref{fig:vlalight_overview}, VLALight first extracts movement-level traffic conditions and temporal dynamics from multi-directional intersection videos. It then routes relevant perceptions from neighbouring intersections to support coordinated network-wide signal control. Our training proceeds in two stages: (1) supervised fine-tuning establishes visual traffic understanding and signal decision-making, and (2) cooperative agentic RL optimizes network-wide coordination using complementary local and network-level traffic rewards. During RL rollouts, the model explores Fast and Slow reasoning modes at each intersection, allowing different intersections in the same network-level rollout to use different modes and learning to balance decision quality and computational cost.

\vspace{-5pt}
\subsection{Vision-based Traffic Signal Control}

\vspace{-5pt}
\paragraph{Spatiotemporal Traffic Perception.}

At each decision step $t$, the visual observation $o_i^t$ contains $K$ frames from the four approaches to intersection $i$. Rather than relying on precomputed traffic states, VLALight directly interprets these multi-frame observations to identify vehicles and their movements and to infer how traffic demand and queues evolve over the observation window. For each traffic lane $l$, the model reasons to summarize the visual dynamics as:
\begin{equation}
    \tilde{o}_{i,l}^t = \left[v_{i,l}^t,\; q_{i,l}^t,\; \Delta v_{i,l}^t,\;\Delta q_{i,l}^t\right]
    =f_{\mathrm{perc}}(o_{i,l}^t),
\end{equation}
where $v_{i,l}^t$ and $q_{i,l}^t$ denote the observed vehicle count and queue length, $\Delta v_{i,l}^t$ and $\Delta q_{i,l}^t$ characterize their changes across the observed frames, and $f_{\mathrm{perc}}$ is the traffic perception process.

However, local perception is insufficient for coordinated control. Vehicles observed at one intersection will travel toward adjacent intersections and alter their near-future demand before they become visible to their views. To capture these neighbouring dynamics, VLALight aggregates movement-level perceptions according to lane connectivity and traffic-flow relations. For each intersection, we collect its perception results and transmit them to adjacent intersections as follows:
\begin{equation}
    c_i^t
    =\left\{\tilde{o}_{l,j}^t\,\middle|\,l\in\mathcal{N}(i),\;j\in\mathcal{L}_{l\rightarrow i}\right\},
\end{equation}
where $\mathcal{N}(i)$ is the set of intersections adjacent to $i$, and $\mathcal{L}_{l\rightarrow i}\subseteq\mathcal{L}_l$ are the lane connections from $l$ to $i$. $c_i^t$ preserves both the temporal dynamics observed at neighbouring intersections and the spatial connectivity, providing topology-aware context for coordinated traffic signal control. The detailed cooperative-context construction is described in Appendix~A.8.

\paragraph{Efficiency-Aware Spatiotemporal Reasoning.}
VLALight integrates the local lane-level perceptions $\tilde{O} = \{\tilde{o}_{i,l}^t\}_{l\in\mathcal{L}_i}$
 with the routed coordination context $c_i^t$ to analyze traffic dynamics and select the signal action that maximizes network-wide traffic efficiency. The VLA model first selects a reasoning mode $m_i^t\in\{\texttt{fast},\texttt{slow}\}$ according to the complexity of the observed traffic. Fast mode directly predicts a signal phase, whereas slow mode generates a structured reasoning sequence $Y_i^t$ to perform deep reasoning on the traffic conditions and candidate phases before decision-making:
\begin{equation}
    \pi_\theta(\tilde{O}_i^t,c_i^t,\mathcal{A}_i)
    =
    \begin{cases}
        (m_i^t,a_i^t),
        & \text{if } m_i^t=\texttt{fast},\\
        (m_i^t,Y_i^t,a_i^t),
        & \text{if } m_i^t=\texttt{slow},
    \end{cases}
    \qquad a_i^t\in\mathcal{A}_i.
\end{equation}
Fast mode directly predicts a phase, whereas Slow mode additionally generates a reasoning sequence before selecting the phase. The resulting action is applied at intersection $i$ during the next control interval, allowing VLALight to allocate additional computation only when deeper reasoning is beneficial. The detailed reasoning template is presented in Appendix~\ref{app:prompt_templates}.
}

\vspace{-5pt}
\siqi{
\subsection{Cooperative Policy Training}

Effective multi-intersection control requires the policy to acquire video-grounded decision capabilities before learning how local phase choices affect neighbouring traffic and network-level outcomes. To establish this initial capability, we perform cold-start supervised fine-tuning from a pretrained vision-language model using video-grounded perception and signal-decision trajectories. To optimize coordination after initialization, we apply cooperative reinforcement learning with local and network-level traffic feedback. To allocate computation according to traffic complexity, we further train the policy with adaptive reasoning to balance decision quality and inference cost.

\vspace{-5pt}
\paragraph{Cold-Start Supervised Training.}
Directly adapting pretrained VLMs to traffic signal control is unreliable because their pretraining rarely covers signal-control scenarios or the spatiotemporal reasoning required for traffic decision-making. We address this issue with a two-stage supervised cold start that separately establishes visual traffic understanding and signal decision-making.
In the first stage, VLALight learns to extract movement-level traffic states from videos. Ground-truth traffic observations are converted into perception targets $\tilde{o}_{i,j}^t$, including vehicle counts, queue lengths, and their temporal variations, and used to supervise the mapping from visual observations to structured traffic perceptions.
In the second stage, we construct reasoning trajectories using a frontier teacher model (e.g., DeepSeek-V4) with access to ground-truth traffic states. Given the perception labels and feasible action space, the teacher generates a reasoning sequence $Y_i^t$ and a signal action $a_i^t$. We then assign the reasoning mode $m_i^t$ based on the queue gap between the two candidate phases with the largest queues: samples with a gap of at least five vehicles are labeled Fast, and the others Slow. For Fast samples, we use the phase with the largest queue as the signal target, which agrees with the teacher-selected phase for most samples, and omit the reasoning target; for Slow samples, we retain the teacher-generated reasoning and signal. The resulting perception--reasoning--decision tuples are then used to jointly fine-tune VLALight with supervised training:
\begin{equation}
\mathcal{L}_{\mathrm{SFT}}=
-
\sum_{\tau\in\mathcal{D}_{\mathrm{SFT}}}
\big[
\log \pi_{\theta}(\tilde{o}_i^t\mid o_i^t)
+
\log \pi_{\theta}(m_i^t,Y_i^t,a_i^t\mid \tilde{o}_i^t,c_i^t,\mathcal{A}_i)
\big],
\end{equation}
where the second term includes the mode, reasoning, and action. $Y_i^t$ is omitted when $m_i^t=\texttt{fast}$.

\vspace{-5pt}
\paragraph{Efficiency-Aware Cooperative VLA RL.}

Building on the supervised initialization, we further optimize VLALight with RL to improve network-wide coordination while controlling reasoning cost. The online training data organization and hyperparameters are summarized in Appendix~\ref{app:training_settings}. To help preserve video-grounded perception during RL, we retain a perception SFT loss and accumulate its gradients with those from decision rollouts before each optimizer step. We consider four optimization signals: local traffic efficiency $r_{i,\mathrm{loc}}^t$, which measures traffic improvement at the controlled intersection; network-wide traffic efficiency $r_{\mathrm{net}}^t$, which captures the impact on surrounding intersections; output-format consistency $r_{\mathrm{fmt}}^t$; and reasoning cost:
\begin{equation}
\kappa_i^t=
\alpha\left[
1-\exp\left(-\frac{n_i^t}{\tau}\right)
\right],
\end{equation}
where $n_i^t$ denotes the number of generated tokens, $\tau$ controls the growth rate, and $\alpha$ bounds the maximum cost. These signals differ substantially in scale and characterize different aspects of the policy. Following the reward-decoupled normalization principle of GDPO~\citep{liu2026gdpogrouprewarddecouplednormalization}, we normalize each reward signal independently within each intersection's rollout group, pooling candidates assigned to both reasoning modes. Specifically, for rollout $k$ and reward signal $d$,
\begin{equation}
A_{k,d}^t=
\frac{r_{k,d}^t-\mu_d^t}
{\sigma_d^t+\epsilon},
\end{equation}
where $\mu_d^t$ and $\sigma_d^t$ are the group-wise mean and standard deviation of $d$, computed across candidates in the rollout group for the corresponding intersection. The reasoning-cost advantage is computed analogously from $\kappa_k^t$. We then construct the overall advantage as:
\begin{equation}
A^t_k=
\lambda_{\mathrm{loc}}A_{k,\mathrm{loc}}^t
+\lambda_{\mathrm{net}}A_{k,\mathrm{net}}^t
+\lambda_{\mathrm{fmt}}A_{k,\mathrm{fmt}}^t
-\lambda_{\mathrm{cost}}A_{k,\mathrm{cost}}^t,
\end{equation}
where $\lambda_d$ ($d\in\{\mathrm{loc},\mathrm{net},\mathrm{fmt},\mathrm{cost}\}$) is the weight of reward signal $d$. This dimension-wise normalization prevents large-scale traffic rewards from overwhelming format or efficiency signals and provides a balanced learning signal for cooperative control and cost-effective reasoning.

However, the unconstrained sampling can lead to imbalanced reasoning-mode exploration, causing one mode to dominate the policy update before the model has learned their relative utility. We therefore explicitly construct balanced fast- and slow-mode rollouts. At each control step $t$, we generate a group of $N$ candidate rollouts from the same visual observation and traffic state, and manually prepend an equal number of \texttt{<mode>fast</mode>} and \texttt{<mode>slow</mode>} tags as fixed generation prefixes for each intersection. Conditioned on the assigned mode, VLALight generates the corresponding reasoning sequence and the signal action. Using the group-normalized rewards, we compare the mean utilities of the Fast and Slow rollouts within each intersection to guide mode selection. The optimization objective is to maximize the group-relative advantage:
\begin{equation}
    \mathcal{L}_{RL}
    =
    -\mathbb{E}_{k}
    \left[
    \min\left(
    \rho_k A_{k},
    \operatorname{clip}(\rho_k,1-\epsilon,1+\epsilon)A_{k}
    \right)
    \right],
    \qquad
    \rho_k=\frac{\pi_{\theta}(y_k\mid x_k)}{\pi_{\theta_{\mathrm{old}}}(y_k\mid x_k)}.
\end{equation}
where $y_k$ is the generated rollout output, $x_k$ is its conditioning context, and $\epsilon$ is the clipping epsilon.
}

\vspace{-10pt}
\section{Experiments}
We evaluate VLALight across traffic networks with diverse scales and demand patterns, and further analyze the contribution of each design component and its adaptive reasoning behavior. Specifically, we investigate the following research questions:

\begin{itemize}[leftmargin=0.5cm,itemsep=0pt,topsep=0.2em,parsep=0pt]
    \item \textbf{RQ1}: How does VLALight perform compared with existing transportation-based, RL-based, and LLM/VLM-based approaches in traffic signal control under diverse traffic scenarios?
    
    \item \textbf{RQ2}: How do cooperative information, the network-level cooperative reward, and balanced Fast/Slow rollouts contribute to traffic control performance beyond supervised fine-tuning?
    
    \item \textbf{RQ3}: How does VLALight allocate Fast and Slow reasoning in response to traffic pressure, and what trade-off does adaptive mode selection achieve among control performance, token usage, and inference latency?
\end{itemize}

\vspace{-5pt}
\subsection{Experimental Setup}

\vspace{-5pt}
\paragraph{Datasets.}
We evaluate VLALight on seven real-world traffic-flow datasets (Jinan 1--3, Hangzhou 1--2, and New York 1--2) from three urban networks~\citep{wei2019survey}. Detailed network configurations and trace statistics are provided in Appendix~\ref{app:datasets}.

\vspace{-5pt}
\paragraph{Environment Settings.}
We use TranSimHub, a video-enabled three-dimensional traffic simulation platform built on the SUMO microscopic traffic simulator~\citep{wang2025transsimhub,lopez2018sumo}, to replay each trace and render the roadside observations. Each episode covers 3,600~s of simulated traffic. At every control step, each controlled intersection receives four directional roadside videos, and the controller selects one feasible phase from the common phase set. The four controlled phases are ETWT (east--west through), ELWL (east--west left-turn), NTST (north--south through), and NLSL (north--south left-turn). Each green phase lasts 25~s, followed by a 5~s yellow transition.

\vspace{-5pt}
\paragraph{Compared Methods.}
We compare VLALight with representative controllers from four groups: transportation-engineering, RL-based, LLM-based, and VLM-based methods. Transportation-engineering and LLM-based controllers directly use complete and accurate traffic states provided by the simulator, whereas VLALight and VLM-based controllers operate on visual observations. Detailed descriptions of the compared methods are provided in Appendix~\ref{app:compared_methods}.

\vspace{-5pt}
\paragraph{Evaluation Metrics.}
We evaluate traffic-control performance using Average Travel Time (ATT), Average Queue Length (AQL), and Average Waiting Time (AWT)~\citep{zhang2022expression}. ATT is the mean elapsed time for vehicles to travel from their origins to their destinations. AQL is the mean number of queued vehicles over the road network during the episode. AWT is the mean time vehicles spend waiting at intersections before completing their movements.

\begin{table*}[t]
    \caption{Traffic-control performance on the Jinan, Hangzhou, and New York datasets. Lower values are better. The best, second-best, and third-best results are highlighted through boldface, double underline, and underline, respectively. E- and A- denote Efficient and Advanced, respectively; Ge-4-31B-IT denotes Gemma-4-31B-IT.}
    \label{tab:regional_results}\centering\scriptsize
    \setlength{\tabcolsep}{1.60pt}
    \renewcommand{\arraystretch}{0.98}
    \resizebox{\textwidth}{!}{%
    \begin{tabular}{@{}l@{\hspace{3pt}}|ccc|ccc|ccc||ccc|ccc||ccc|ccc}
    \toprule
    \multirow{2}{*}{Method} & \multicolumn{9}{c||}{Jinan} & \multicolumn{6}{c||}{Hangzhou} & \multicolumn{6}{c}{New York}\\
     & \multicolumn{3}{c|}{1} & \multicolumn{3}{c|}{2} & \multicolumn{3}{c||}{3} & \multicolumn{3}{c|}{1} & \multicolumn{3}{c||}{2} & \multicolumn{3}{c|}{1} & \multicolumn{3}{c}{2}\\
     & ATT & AQL & AWT & ATT & AQL & AWT & ATT & AQL & AWT & ATT & AQL & AWT & ATT & AQL & AWT & ATT & AQL & AWT & ATT & AQL & AWT\\
    \midrule
    \multicolumn{22}{c}{Transportation-Engineering Methods}\\
    FixedTime & 458.95 & 383.91 & 237.36 & 366.59 & 185.60 & 152.92 & 394.99 & 206.57 & 136.14 & 544.46 & 207.24 & 269.66 & 464.08 & 274.97 & 229.64 & 1464.24 & 2716.68 & 1171.09 & 1666.71 & 3940.70 & 1387.68\\
    MaxPressure & 270.97 & 163.77 & 79.79 & 267.68 & 109.82 & 74.35 & 260.02 & 135.30 & 73.35 & 306.45 & 65.22 & 63.13 & 304.98 & 120.88 & 73.96 & 1215.43 & 2395.16 & 894.25 & 1467.47 & 4066.03 & 1202.50\\
    \midrule
    \multicolumn{22}{c}{RL-Based Methods}\\
    PressLight & 286.26 & 189.89 & 94.79 & 281.39 & 125.44 & 87.26 & 267.93 & 147.34 & 82.03 & 350.22 & 98.71 & 104.79 & 345.19 & 183.61 & 120.24 & 1511.56 & 3285.03 & 1322.43 & 1687.35 & 4661.39 & 1521.07\\
    MPLight & 330.62 & 286.63 & 153.34 & 281.04 & 125.97 & 89.15 & 273.92 & 157.63 & 89.65 & 317.27 & 73.72 & 75.80 & 311.38 & 133.37 & 84.39 & 1229.75 & 2590.86 & 999.76 & 1526.80 & 4198.63 & 1303.49\\
    CoLight & 272.15 & 166.55 & 81.70 & 267.35 & 108.98 & 75.03 & 262.32 & 138.57 & 76.43 & 308.80 & 66.23 & 66.64 & 307.51 & 126.48 & 79.17 & \underline{1032.32} & 2056.00 & 734.90 & \underline{\underline{1256.04}} & \underline{3339.83} & \underline{970.57}\\
    E-MPLight & 293.79 & 204.88 & 103.39 & 349.17 & 201.84 & 152.36 & 312.26 & 215.90 & 127.38 & 384.44 & 132.26 & 145.23 & 456.64 & 342.78 & 250.48 & 1322.91 & 2665.49 & 1053.87 & 1638.79 & 4667.94 & 1426.80\\
    E-CoLight & 273.99 & 169.18 & 82.40 & 265.66 & 107.91 & 72.17 & 259.71 & 135.48 & 73.14 & 306.85 & 64.88 & 62.95 & 306.35 & 122.76 & 76.42 & 1104.93 & 2183.63 & 784.81 & 1307.53 & 3521.47 & 1032.81\\
    A-CoLight & 269.57 & 162.60 & 78.78 & 267.60 & 111.03 & 75.06 & 258.81 & 134.08 & 72.60 & 302.51 & 61.15 & 59.66 & 304.91 & 120.62 & 75.04 & 1094.28 & 2214.17 & 807.06 & 1303.36 & 3590.56 & 1051.77\\
    CityLight & 364.87 & 317.15 & 171.01 & 322.59 & 171.63 & 127.74 & 321.50 & 222.25 & 133.37 & 360.27 & 107.85 & 115.80 & 528.86 & 389.34 & 310.30 & 1304.88 & 2685.88 & 1037.71 & 1525.28 & 4181.28 & 1290.92\\
    \midrule
    \multicolumn{22}{c}{LLM-Based Methods}\\
    LLMLight & \textbf{263.90} & \underline{\underline{154.86}} & \underline{\underline{73.80}} & \underline{\underline{262.79}} & \underline{104.48} & \underline{69.12} & \textbf{254.46} & \underline{128.56} & \underline{\underline{69.05}} & \underline{299.53} & \underline{58.86} & \underline{56.23} & \underline{296.39} & \underline{108.58} & \underline{65.81} & 1051.99 & \underline{1908.08} & \underline{682.86} & 1347.41 & 3647.39 & 1032.21\\
    CoLLMLight & \underline{\underline{265.28}} & \underline{155.89} & \underline{74.97} & \textbf{262.38} & \underline{\underline{103.28}} & \underline{\underline{68.98}} & \underline{\underline{255.22}} & \underline{\underline{128.41}} & \underline{69.49} & \underline{\underline{295.48}} & \textbf{54.93} & \underline{\underline{52.36}} & \textbf{294.49} & \underline{\underline{104.88}} & \underline{\underline{64.00}} & \underline{\underline{1009.34}} & \underline{\underline{1798.50}} & \underline{\underline{644.86}} & \underline{1263.85} & \underline{\underline{3304.41}} & \underline{\underline{943.32}}\\
    \midrule
    \multicolumn{22}{c}{VLM-Based Methods}\\
    VLMLight & 381.64 & 368.59 & 196.32 & 295.38 & 140.70 & 99.90 & 415.10 & 567.78 & 370.77 & 310.73 & 64.44 & 61.22 & 318.51 & 140.60 & 95.84 & 1271.80 & 2610.89 & 993.17 & 1743.75 & 4928.48 & 1526.34\\
    Qwen3.5-27B & 720.28 & 930.73 & 551.68 & 735.28 & 698.37 & 562.24 & 730.64 & 843.04 & 566.67 & 527.36 & 252.07 & 292.45 & 464.15 & 363.76 & 254.68 & 1620.75 & 3342.57 & 1397.52 & 1782.92 & 4943.55 & 1655.35\\
    Qwen3.5-9B & 929.73 & 1216.28 & 754.35 & 891.58 & 870.12 & 709.09 & 786.08 & 952.59 & 615.97 & 767.84 & 463.15 & 550.13 & 615.56 & 572.92 & 416.91 & 1770.33 & 3781.35 & 1665.94 & 1909.77 & 5278.27 & 1816.58\\
    Ge-4-31B-IT & 680.47 & 875.06 & 510.09 & 684.15 & 633.87 & 508.59 & 660.66 & 752.95 & 492.89 & 453.02 & 191.90 & 219.01 & 421.09 & 294.93 & 205.24 & 1592.13 & 3409.36 & 1367.92 & 1766.68 & 3409.36 & 1367.92\\
    VLALight & \underline{267.05} & \textbf{154.74} & \textbf{73.31} & \underline{265.04} & \textbf{103.01} & \textbf{67.48} & \underline{256.98} & \textbf{128.38} & \textbf{67.77} & \textbf{294.90} & \underline{\underline{55.94}} & \textbf{52.05} & \underline{\underline{296.29}} & \textbf{99.03} & \textbf{60.91} & \textbf{975.80} & \textbf{1565.44} & \textbf{566.82} & \textbf{1240.11} & \textbf{3121.64} & \textbf{880.49}\\
    \bottomrule
    \end{tabular}
    }%
\vspace{-10pt}
\end{table*}

\vspace{-5pt}
\subsection{Overall Performance (RQ1)}
Table~\ref{tab:regional_results} compares traffic-control performance across seven datasets from Jinan, Hangzhou, and New York. The baselines reveal a clear progression. MaxPressure improves substantially over FixedTime by responding to traffic pressure. CoLight generally surpasses non-cooperative RL controllers by exchanging information across intersections. LLMLight and CoLLMLight further benefit from enhanced reasoning capabilities over structured traffic states. By contrast, off-the-shelf VLMs perform poorly, indicating that general visual-language capabilities alone are insufficient for traffic-specific spatiotemporal reasoning. Despite being optimized with online RL only on Jinan 1 and Hangzhou 1, VLALight successfully generalizes to all five unseen traffic traces.

VLALight achieves the best result in 16 of 21 dataset--metric comparisons despite using limited-view roadside videos, whereas transportation-engineering and LLM baselines receive complete simulator states. Its advantage is significant on the two 196-intersection New York networks: it ranks first in ATT, AQL, and AWT on two datasets. This validates our cooperative multi-intersection RL design, which integrates topology-routed neighbouring perceptions with local and network-level rewards, enabling each agent to optimize immediate traffic conditions while accounting for downstream effects across the network. Compared with state-of-the-art baselines, VLALight reduces AQL and AWT by 13.0\% and 12.1\% on New York 1, and by 5.5\% and 6.7\% on New York 2. These results demonstrate effective network-wide coordination from partial visual observations.

\begin{figure*}[t]
    \noindent
    \begin{minipage}[t]{0.56\textwidth}
    \vspace{0pt}
    \centering\scriptsize
    \captionof{table}{Ablation results on New York 1 and New York 2. ``w/o coop.'' denotes removal of cooperative information, ``w/o balance'' denotes removal of balanced rollouts, and ``w/o net. reward'' denotes removal of the network-level cooperative reward.}
    \label{tab:ablation_results}
    \setlength{\tabcolsep}{3.0pt}
    \renewcommand{\arraystretch}{0.82}
    \resizebox{\linewidth}{!}{%
    \begin{tabular}{@{}lccc|ccc@{}}
    \toprule
    \multicolumn{1}{l}{\scriptsize Variant} & \multicolumn{3}{c}{New York 1} & \multicolumn{3}{c}{New York 2}\\
    \cmidrule(lr){2-4}\cmidrule(lr){5-7}
     & ATT & AQL & AWT & ATT & AQL & AWT\\
    \midrule
    SFT & 1171.08 & 2439.95 & 866.98 & 1439.73 & 4021.12 & 1164.74\\
    w/o coop. & 1067.60 & 2365.90 & 914.98 & 1317.63 & 3690.11 & 1113.65\\
    w/o balance & 984.76 & 1654.22 & 595.02 & 1296.10 & 3328.31 & 938.46\\
    w/o net. reward & 991.34 & 1647.22 & 599.61 & 1261.11 & 3230.98 & 924.42\\
    VLALight & \textbf{975.80} & \textbf{1565.44} & \textbf{566.82} & \textbf{1240.11} & \textbf{3121.64} & \textbf{880.49}\\
    \bottomrule
    \end{tabular}
    }%
    \end{minipage}\hfill
    \begin{minipage}[t]{0.40\textwidth}
    \vspace{0.18em}
    \centering
    \includegraphics[width=\linewidth,height=1.28in,keepaspectratio]{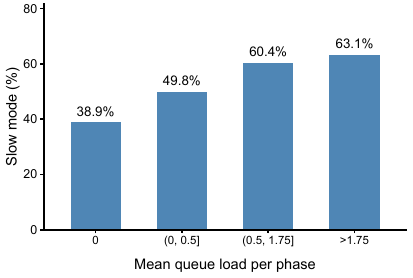}
    \vspace{0.55em}
    {\scriptsize
    \captionof{figure}{Slow-mode ratio across local queue-load groups.}\label{fig:pressure_mode}
    }
    \end{minipage}
\vspace{-10pt}
\end{figure*}

\vspace{-5pt}
\subsection{Ablation Studies (RQ2)}

Table~\ref{tab:ablation_results} evaluates cooperative information, network-level reward, and balanced Fast/Slow rollouts on both New York datasets. VLALight is initialized from Qwen3.5-4B. We exclude the original Qwen3.5-4B model because it cannot output structurally valid control actions. The post-trained VLALight consistently outperforms SFT across all metrics, demonstrating that online cooperative RL provides substantial gains beyond supervised initialization. Among the RL ablations, removing cooperative information causes the largest degradation, particularly in AQL and AWT. This result indicates that topology-routed neighbouring perceptions compensate for the limited field of view at each intersection and are essential for modeling network-wide traffic interactions. Removing the network-level reward also degrades every metric, showing that optimizing local traffic alone does not reliably produce network-wide improvements. Finally, removing balanced Fast/Slow rollouts reduces performance, especially on New York 2, indicating that explicit exposure to both reasoning modes improves exploration and policy learning. Together, these results validate the effectiveness of our proposed training procedure in VLALight.

\subsection{Adaptive Reasoning and Inference Efficiency (RQ3)}

\begin{table*}[t]
\caption{Adaptive reasoning token statistics and ATT on New York datasets. Fast and Slow report the percentages of decisions assigned to each mode.}
\label{tab:reasoning_results}\centering\scriptsize
\setlength{\tabcolsep}{3.2pt}
\begin{tabular}{@{}llccccc@{}}
\toprule
\multirow{2}{*}{Trace} & \multirow{2}{*}{Method} & \multicolumn{4}{c}{Reasoning statistics} & \multicolumn{1}{c}{Control performance}\\
\cmidrule(lr){3-6}\cmidrule(l){7-7}
 & & Fast (\%) & Slow (\%) & Avg. Tokens (Slow) & Min--max Tokens (Slow) & ATT\\
\midrule
New York 1 & VLALight  & 51.92 & 48.08 & 220.47 & 54--545   & \textbf{975.80}\\
New York 1 & w/o balanced rollouts & 27.56 & 72.44 & 209.42 & 62--649   & 984.76\\
New York 1 & SFT       & 59.72 & 40.28 & 233.15 & 62--1,230 & 1171.08\\
\addlinespace
New York 2 & VLALight  & 51.24 & 48.76 & 208.26 & 58--682   & \textbf{1240.11}\\
New York 2 & w/o balanced rollouts & 28.38 & 71.62 & 204.92 & 62--609   & 1296.10\\
New York 2 & SFT       & 60.27 & 39.73 & 230.50 & 62--1,124 & 1439.73\\
\bottomrule
\end{tabular}
\end{table*}

\paragraph{Adaptive Mode Allocation.}
Table~\ref{tab:reasoning_results} examines how balanced rollout training affects inference-time mode allocation, slow-mode token usage, and control performance on the two New York datasets. Balanced rollout training produces a markedly different allocation policy. VLALight invokes slow reasoning for approximately 48\% of decisions on both traces, whereas the variant without balanced rollouts selects slow-mode more than 72\% of the time but still yields worse ATT. Thus, simply allowing the policy to choose a mode does not produce an effective computation strategy. Without balanced Fast/Slow exploration, training becomes biased toward the more expensive mode without a corresponding control benefit. SFT shows the opposite behavior: it invokes slow mode less frequently, but produces considerably longer slow-mode outputs and substantially weaker control performance. This indicates that our balanced rollout exploration enables VLALight to learn when additional deliberation is genuinely beneficial. As a result, the policy achieves a more cost-effective trade-off between control utility and reasoning overhead.

\begin{table}[t]
\centering
\begin{minipage}[t]{0.48\columnwidth}
    \vspace{0pt}
    \centering
    {\scriptsize
    \setlength{\tabcolsep}{1.35pt}
    \renewcommand{\arraystretch}{0.90}
    \caption{Control performance under forced and adaptive mode allocation.}
    \label{tab:forced_mode}
    \vspace{0.45em}
    \begin{tabular}{lrrrrrr}
    \toprule
    & \multicolumn{3}{c}{\textit{Jinan 1}} & \multicolumn{3}{c}{\textit{Hangzhou 2}}\\
    \cmidrule(lr){2-4}\cmidrule(lr){5-7}
    Mode & ATT & AQL & AWT & ATT & AQL & AWT\\
    \midrule
    Pure Fast & 268.93 & 157.75 & 74.88 & 297.95 & 101.22 & 62.86\\
    Pure Slow & \textbf{263.48} & 154.97 & \textbf{72.86} & \textbf{295.28} & \textbf{97.09} & \textbf{59.90}\\
    VLALight & 267.05 & \textbf{154.74} & 73.31 & 296.29 & 99.03 & 60.91\\
    \bottomrule
    \end{tabular}\\[-1pt]
    }
\end{minipage}\hfill
\begin{minipage}[t]{0.48\columnwidth}
    \vspace{0pt}
    \centering
    {\scriptsize
    \setlength{\tabcolsep}{1.35pt}
    \renewcommand{\arraystretch}{0.90}
    \caption{Average per-decision inference latency on two H100 GPUs over 500 sampled intersection decisions (seconds).}
    \label{tab:inference_latency}
    \begin{tabular}{lrccr}
    \toprule
    Mode & Samples & Perception time & Decision time & Total \\
    \midrule
    Pure Fast & 174 & 2.601 & 0.265 & \textbf{2.866} \\
    Pure Slow & 326 & 2.563 & 1.944 & \textbf{4.507} \\
    Overall & 500 & 2.576 & 1.360 & \textbf{3.936} \\
    \bottomrule
    \end{tabular}
    }
\end{minipage}
\vspace{-10pt}
\end{table}

\paragraph{Control Quality and Inference Efficiency.}
To isolate the effect of reasoning depth, we fix the mode token at inference to construct Pure Fast and Pure Slow policies and compare them with VLALight's adaptive allocation in Table~\ref{tab:forced_mode}. We further profile per-decision latency on two H100 GPUs over 500 sampled decisions; Table~\ref{tab:inference_latency} reports the mean perception, decision, and total latency for Fast and Slow decisions and for the adaptive policy overall.
Pure Slow achieves the best result on most metrics, confirming that extended reasoning benefits difficult control decisions, whereas Pure Fast is consistently weaker. Importantly, VLALight remains close to Pure Slow across both datasets and even achieves the lowest AQL on Jinan 1, showing that adaptive mode selection preserves most of the benefit of deeper reasoning without paying its cost for every decision. Fast and slow decisions require 2.866~s and 4.507~s on average, respectively. VLALight's adaptive allocation reduces the overall mean to 3.936~s, 12.7\% below the slow-mode latency. Perception time is nearly constant across modes, while decision time accounts for the efficiency difference, directly validating the contribution of adaptive reasoning.

\subsubsection{Traffic Pressure and Adaptive Mode Selection}
We next examine whether the learned allocation responds meaningfully to traffic complexity. Using video-grounded perception records, we group decisions by the average queue load across the four candidate phases: queue-free observations and three positive-load groups separated at 0.5 and 1.75 vehicles per phase. As shown in Figure~\ref{fig:pressure_mode}, the Slow-mode ratio rises monotonically from $38.86\%$ in queue-free conditions to $49.81\%$, $60.39\%$, and $63.14\%$ as queue pressure increases. This trend provides direct evidence that VLALight reserves additional computation for more demanding traffic states rather than selecting modes arbitrarily. The representative cases in Figure~\ref{fig:pressure_mode_cases} further show that fast reasoning is used when one local phase is clearly preferable, while slow reasoning is activated when competing phases or topology-routed neighbour pressure make the decision more ambiguous.
\input{figures/traffic_pressure_mode_cases.tex}

\vspace{-5pt}
\section{Related Work}

\vspace{-5pt}
\paragraph{Traffic Signal Control.}

Traffic signal control (TSC) has evolved from fixed schedules and pressure-based rules to RL controllers that learn phase policies and coordinate neighbouring intersections through attention, graphs, and multi-agent communication~\citep{koonce2008timing,hunt1982scoot,varaiya2013maxpressure,oroojlooy2020attendlight,wu2023transformerlight,devailly2022igrl,liang2022oam,lou2022metarl,wei2018intellilight,zheng2019frap,wei2019presslight,chen2020mplight,wei2019colight,wu2021dynstgat,yu2020macar,ruan2024coslight,mercader2020maxpressure}. LLM-assisted methods then introduced tool-based human-mimetic reasoning and arterial coordination, while iLLM-TSC uses an LLM to revise RL decisions under incomplete observations and rare events~\citep{wang2024llmassisted,tang2024arterial,pang2026illmtsc}. LLMLight treats the LLM as a direct controller; CoLLMLight adds asynchronous network-wide cooperation; and Traffic-R1 applies reinforcement learning to improve reasoning and generalization~\citep{lai2025llmlight,yuan2026collmlight,zou2026trafficr1}. VLMLight incorporates visual meta-control but delegates routine actions to RL~\citep{wang2025vlmlight}. Consequently, prior systems largely depend on structured simulator states or modular planners, which leaves end-to-end TSC from physical visual observations unexplored.

\vspace{-5pt}
\paragraph{Vision-Language-Action Models.}

Vision-language-action (VLA) models directly map visual-language inputs to executable actions. RT-2 introduced action tokenization to transfer web-scale knowledge to robotic control, while OpenVLA provided an open generalist manipulation model~\citep{brohan2023rt2,kim2024openvla}. Embodied Chain-of-Thought subsequently grounded intermediate reasoning in plans, objects, and robot states~\citep{zawalski2024ecot}. Efficiency-oriented work explores state-space backbones, compact policies, frequency-space action tokenization, training-free compression, and adaptive token caching~\citep{liu2024robomamba,wen2025tinyvla,pertsch2025fast,yang2025efficientvla,xu2025vlacache}. AutoVLA extends unified action generation to autonomous driving, using supervised Fast/Slow modes and reinforcement fine-tuning to balance planning quality and reasoning cost~\citep{zhou2025autovla}.

\vspace{-5pt}
\section{Conclusion}
We introduced VLALight, the first VLA framework for end-to-end traffic signal control from multi-view roadside videos. VLALight directly connects physical visual observations with executable signal actions through multi-target spatiotemporal reasoning and topology-aware cooperative perception. Its two-stage supervised initialization and cooperative agentic RL enable network-level control with adaptive Fast/Slow reasoning that balances traffic efficiency and inference cost. Experiments on seven real-world traffic-flow datasets across Jinan, Hangzhou, and New York demonstrate consistent improvements over transportation-, RL-, and LLM/VLM-based controllers, while ablations validate the contributions of cooperative perception, network optimization, and adaptive reasoning. These results highlight the potential of VLA models for real-world physical traffic control.

\clearpage
\bibliography{iclr2027_conference}
\bibliographystyle{iclr2027_conference}

\clearpage
\appendix
\section{APPENDIX}
\subsection{Settings of Traffic Signal Control}
\label{app:traffic_control_settings}

\begin{figure}[H]
    \centering
    \begin{tabular}{@{}c@{\hspace{0.015\linewidth}}c@{\hspace{0.015\linewidth}}c@{}}
        \includegraphics[height=4cm, valign=c]{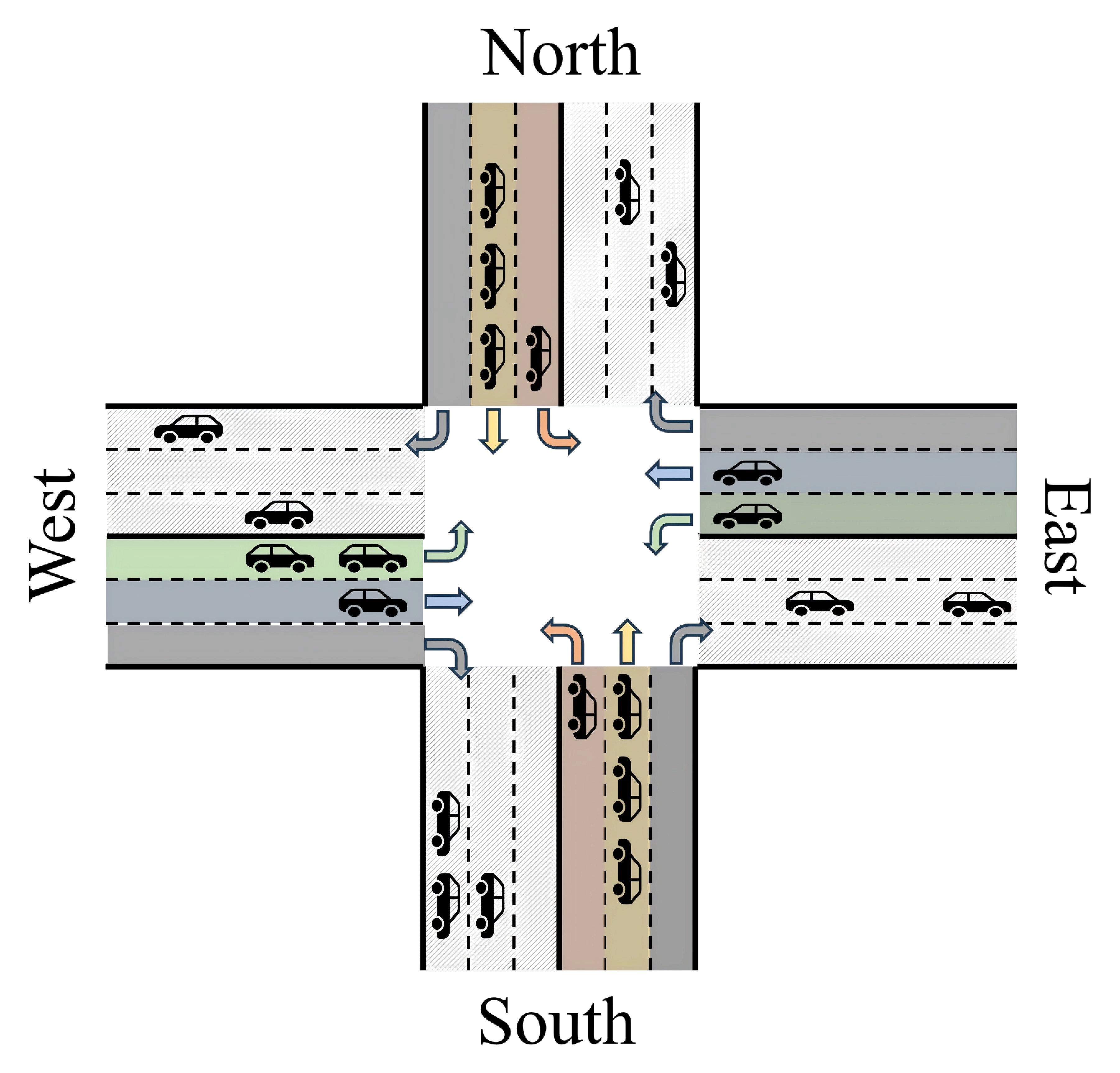} &
        \includegraphics[height=2cm, valign=c]{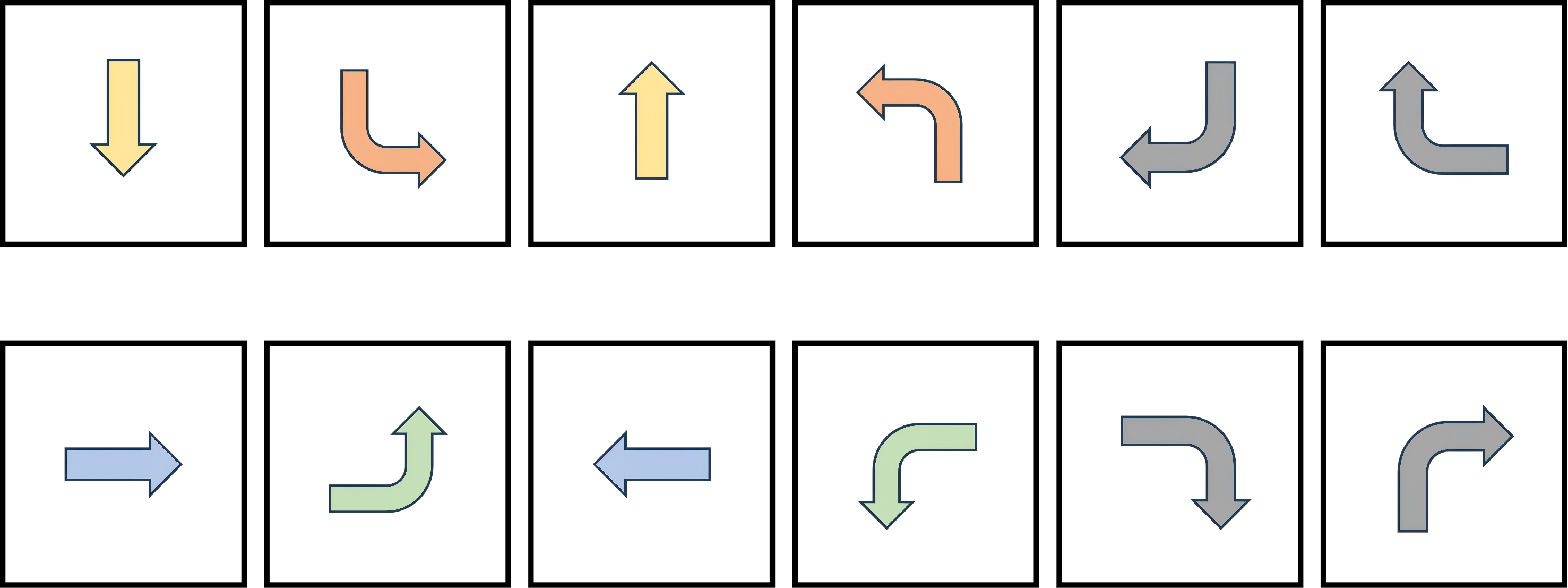} &
        \includegraphics[height=4cm, valign=c]{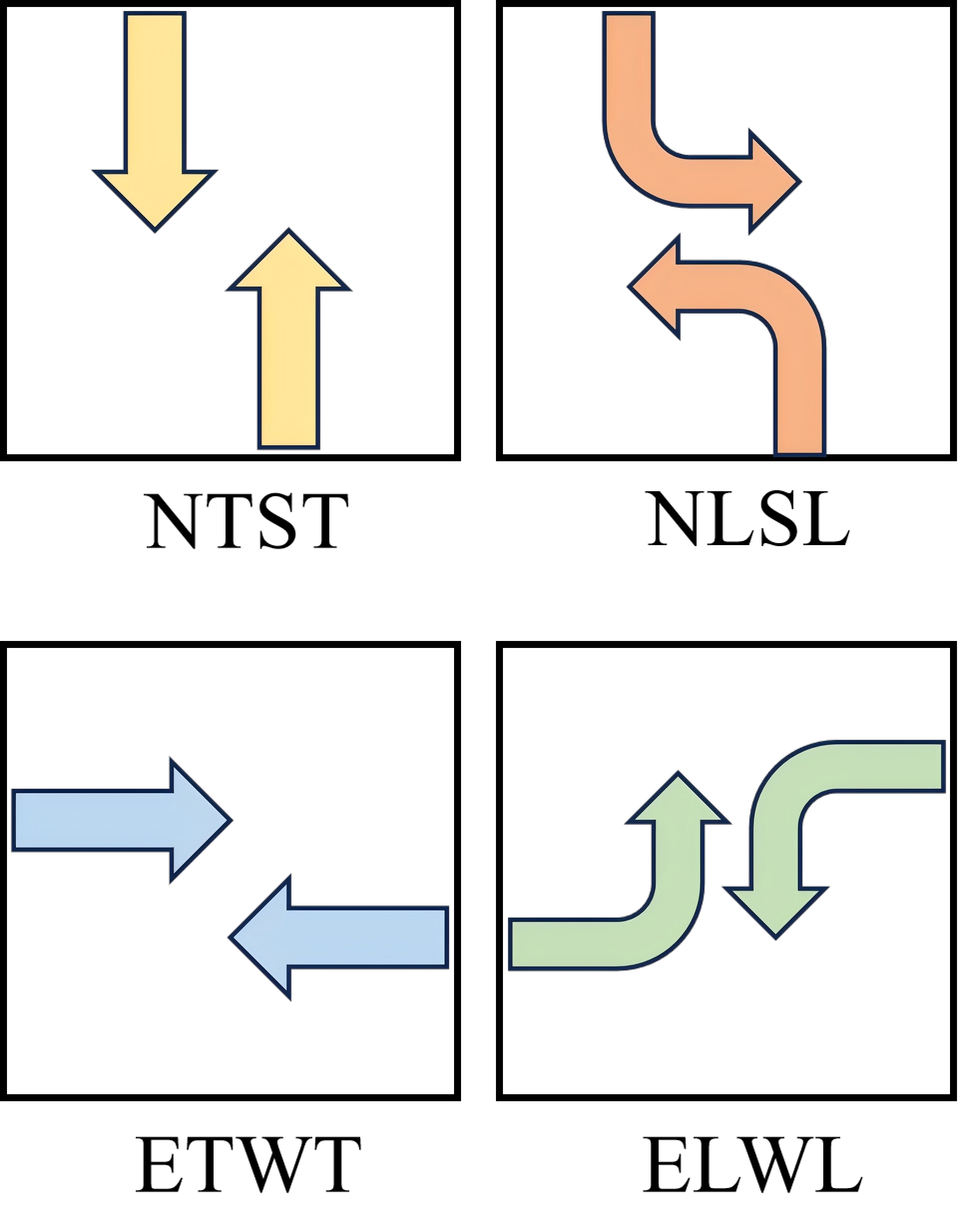} \\
        \textbf{(A) Intersection} &
        \textbf{(B) Twelve Lanes} &
        \textbf{(C) Four Signal Phases}
    \end{tabular}
    \caption{Traffic-signal-control setting used in our experiments. (A) A four-arm intersection. (B) Twelve incoming lanes and their permitted movements; right turns remain permissive. (C) Four controlled signal phases, each pairing non-conflicting movements: ETWT and NTST serve east--west and north--south through movements, respectively, whereas ELWL and NLSL serve the corresponding left-turn movements.}
    \label{fig:traffic_control_setting}
\end{figure}

The benchmark environments share the intersection geometry and phase design shown in Figure~\ref{fig:traffic_control_setting}. Each four-arm intersection contains 12 incoming lanes. Four controlled phases serve through and left-turn movements, while right-turn movements remain permissive.

\subsection{Datasets and Traffic Networks}
\label{app:datasets}

We evaluate VLALight on seven real-world traffic-flow datasets from three urban networks. Each trace specifies vehicle departures and routes for a one-hour episode on its corresponding road network.
\begin{itemize}
    \item \textbf{Jinan}: Located in the Dongfeng sub-district, the network contains 12 intersections arranged as a $3\times4$ grid and three datasets from different collection periods. Each intersection has two 400-m east--west roads and two 800-m north--south roads.
    \item \textbf{Hangzhou}: Located in the Gudang sub-district, the network contains 16 intersections arranged as a $4\times4$ grid and two datasets from different collection periods. Each intersection has two 800-m east--west roads and two 600-m north--south roads.
    \item \textbf{New York}: Covering Manhattan's Upper East Side, the network contains 196 intersections arranged as a $28\times7$ grid. Its two large-scale datasets are derived from taxi-trip demand from different periods, with 300-m road segments.
\end{itemize}
Table~\ref{tab:dataset_statistics} reports the number of vehicles and five-minute arrival statistics for every trace. Arrival statistics are computed from vehicle entry times using 300-s bins; $\mu\pm\sigma$ denotes the mean and population standard deviation across bins, and the final column gives the observed minimum and maximum.
\begin{table}[H]
\caption{Traffic-flow trace statistics. Arrival rates are vehicles per 5 min; values are reported as mean $\pm$ standard deviation, followed by the observed range.}
\label{tab:dataset_statistics}
\centering
\small
\begin{tabular}{l c c r c c}
\toprule
City & Trace & Grid & Vehicles & 5-min arrivals ($\mu\pm\sigma$) & Min--max\\
\midrule
Jinan & 1 & $3\times4$ & 6295 & $524.58\pm98.53$ & 256--672\\
Jinan & 2 & $3\times4$ & 4365 & $363.75\pm74.66$ & 237--493\\
Jinan & 3 & $3\times4$ & 5494 & $457.83\pm46.22$ & 363--544\\
\midrule
Hangzhou & 1 & $4\times4$ & 2983 & $248.58\pm40.45$ & 212--333\\
Hangzhou & 2 & $4\times4$ & 6984 & $582.00\pm318.52$ & 203--1146\\
\midrule
New York & 1 & $28\times7$ & 11058 & $850.62\pm174.20$ & 383--965\\
New York & 2 & $28\times7$ & 16337 & $1256.69\pm264.96$ & 476--1441\\
\bottomrule
\end{tabular}
\end{table}

\subsection{Compared Methods}
\label{app:compared_methods}

This section briefly summarizes the control interfaces and design principles of the methods included in our comparison. The descriptions refer to the configurations used in our experiments.
\begin{description}
\item[FixedTime~\citep{koonce2008timing}:] Applies a predetermined cycle and phase schedule, without adapting the signal plan to the current traffic state.

\item[MaxPressure~\citep{varaiya2013maxpressure}:] Selects the phase with the largest pressure differential, computed from the queue imbalance between incoming and outgoing movements.

\item[PressLight~\citep{wei2019presslight}:] Learns a deep policy whose observation and optimization signal are derived from intersection pressure, thereby adapting the max-pressure principle through reinforcement learning.

\item[MPLight~\citep{chen2020mplight}:] Uses pressure-based traffic features for both state representation and reward design within a multi-agent learning framework built on FRAP-style phase relations.

\item[CoLight~\citep{wei2019colight}:] Models the road network as interacting agents and uses graph attention to exchange information between neighbouring intersections.

\item[Efficient-MPLight and Efficient-CoLight~\citep{wu2021efficientpressure}:] Reduce the state complexity of pressure-based control by using compact pressure observations, while retaining the single-intersection and cooperative variants, respectively.

\item[Advanced-CoLight~\citep{zhang2022expression}:] Extends graph-based cooperative control with richer traffic-state features, including pressure and effective running-vehicle information.

\item[CityLight~\citep{zeng2025citylight}:] Uses parameter-sharing MAPPO to learn a universal policy for heterogeneous intersections, with dedicated modules for encoding and aggregating neighbour influence in city-scale networks.

\item[LLMLight~\citep{lai2025llmlight}:] Uses a large language model as the signal-control agent. Given structured traffic descriptions, the model reasons over candidate phases and produces the control decision.

\item[CoLLMLight~\citep{yuan2026collmlight}:] Extends LLM-based signal control to network-wide coordination by allowing agents at connected intersections to use cooperative traffic information.

\item[VLMLight~\citep{wang2025vlmlight}:] Uses vision-language meta-control and dual-branch reasoning to interpret visual traffic observations and select signal-control actions. In our experiments, we adapt VLMLight by using Qwen3.5-27B to analyze four directional images and route control according to traffic pressure. Under normal conditions, a shared Advanced-CoLight policy selects the phase; high pressure activates VLM-based phase selection and validation. Unlike the original emergency-vehicle trigger, our implementation uses traffic pressure to activate this branch.

\end{description}

\subsection{Model Settings}
\label{app:model_settings}

All RL baselines are trained with shared hyperparameters, including a learning rate of $1\times10^{-3}$, a replay-buffer capacity of 12,000, a sample size of 3,000, and a hidden size of 20. VLALight is built upon Qwen3.5-4B and supervised fine-tuned with LoRA using rank 16, scaling factor 32, dropout 0.05, and a learning rate of $1\times10^{-4}$. Each video input is represented by six frames at a resolution of $512\times960$ pixels.

\subsection{Agentic RL Settings}
\label{app:training_settings}

Starting from the cold-start Qwen3.5-4B policy, we perform agentic reinforcement learning with online SUMO streams from the Jinan 1 and Hangzhou 1 datasets. At each training step, we use two Jinan 1 and two Hangzhou 1 network instances with different random seeds. All intersections within each network instance jointly execute one 30-s control-cycle transition, yielding intersection-level decision records from the synchronized network snapshot. Each decision uses six synchronized candidates, with three Fast and three Slow rollouts. Network-level reward is evaluated for each complete network instance over three subsequent control cycles. The resulting records are updated intersection by intersection with a mini-batch size of 4, for 50 training steps, using a policy learning rate of $1\times10^{-6}$. After independent within-group normalization, the network, local, reasoning-cost, and format dimensions are combined with weights 0.5, 1.0, 0.5, and 1.0, respectively. During these updates, the perception supervised loss is retained with coefficient 0.1 to preserve visual traffic understanding while optimizing the cooperative decision policy.

\subsection{Cooperative VLA Rollout and Policy Update}
\label{app:training_algorithm}

Algorithm~\ref{alg:vlalight_training} summarizes the training procedure used by VLALight. Let $\mathcal{I}$ denote the controlled intersections. For each network decision snapshot, we collect the joint visual observations $\mathbf{V}^{t}=\{V_i^t\}_{i\in\mathcal{I}}$ and generate the corresponding local perceptions and routed cooperative contexts once, obtaining $\mathbf{P}^{t}$ and $\mathbf{C}^{t}$. These inputs are shared by all candidates in the rollout group. A rollout is defined at the network level: the $b$-th candidate jointly assigns a reasoning mode, reasoning sequence, and signal action to every intersection,
\begin{equation}
    \mathbf{m}^{(b),t}=\{m_i^{(b),t}\}_{i\in\mathcal{I}},\qquad
    \mathbf{a}^{(b),t}=\{a_i^{(b),t}\}_{i\in\mathcal{I}},
\end{equation}
and produces one network trajectory $\tau^{(b)}=(s^t,\mathbf{a}^{(b),t},s^{(b),t+1},\ldots,s^{(b),t+H})$. All actions in a candidate are applied jointly before the environment is advanced. The resulting network-level reward is shared by all intersection records belonging to that candidate, whereas local traffic, format, and reasoning-cost signals remain intersection-specific. Let $q_i^t$ denote the queue at intersection $i$, and let $Q^t=\sum_{i\in\mathcal{I}}q_i^t$ and $Q^{(b),t+H}=\sum_{i\in\mathcal{I}}q_i^{(b),t+H}$ denote the total network queues at the initial and terminal states. We define the normalized queue-improvement scores as
\begin{equation}
\label{eq:queue_reward_scores}
\begin{aligned}
r_{\mathrm{net}}^{(b)}&=Q^t-Q^{(b),t+H},\\
r_{\mathrm{local},i}^{(b)}&=q_i^t-q_i^{(b),t+1}.
\end{aligned}
\end{equation}
Thus, the network reward measures the queue reduction from $t$ to the rollout endpoint after three transitions, while the local reward measures the first-transition reduction from $t$ to $t+1$. Positive values indicate queue reduction. The format penalty is assigned as
\begin{equation}
\label{eq:format_penalty}
r_{\mathrm{fmt},i}^{(b)}=\begin{cases}-1,&\text{if the signal is invalid},\\
-\frac{1}{2},&\text{if the signal is valid but the response format is invalid},\\
0,&\text{if the response is valid}.
\end{cases}
\end{equation}
The optimization then uses two complementary updates: a mode-selection update that compares Fast and Slow candidates separately, and a joint policy update that aggregates the rollout signals across all controlled intersections without splitting by reasoning mode.

\begin{algorithm}[H]
\caption{Cooperative VLA Rollout}
\label{alg:vlalight_training}
\begin{algorithmic}[1]
\REQUIRE Joint network observations $\mathbf{V}^t$, road network $\mathcal{G}$, policy $\pi_\theta$, rollout number $N$
\STATE Generate $\mathbf{P}^t$ and routed contexts $\mathbf{C}^t=\mathcal{R}(\mathbf{P}^t,\mathcal{G})$ once from the network snapshot
\STATE Share $(\mathbf{P}^t,\mathbf{C}^t)$ across the $N$ candidate rollouts
\STATE Construct $N$ balanced mode maps $\{\mathbf{m}^{(b),t}\}_{b=1}^{N}$, with Fast and Slow balanced separately for each intersection
\FOR{each rollout $b=1,\ldots,N$}
    \STATE Generate $\{Y_i^{(b),t},a_i^{(b),t}\}_{i\in\mathcal{I}}$ conditioned on $\mathbf{m}^{(b),t}$ and $(\mathbf{P}^t,\mathbf{C}^t)$
    \STATE Apply the joint action $\mathbf{a}^{(b),t}$ and advance the network for $H=3$ transitions; use temperature-zero continuation decisions
    \STATE Compute one network reward $r_{\mathrm{net}}^{(b)}$ and intersection-specific local, format, and cost signals
\ENDFOR
\STATE Broadcast $r_{\mathrm{net}}^{(b)}$ to every intersection in rollout $b$
\STATE Compare the Fast and Slow candidates for each intersection and update the mode selection
\STATE Update the shared policy with the joint rollout signals from all controlled intersections, without splitting the update by mode
\STATE \RETURN updated policy $\pi_\theta$
\end{algorithmic}
\end{algorithm}

\subsection{Two-Stage Cooperative VLA Inference}
\label{app:inference_algorithm}

At deployment, the video agent extracts structured traffic states from the four
approach videos and routes information through the road topology to form
cooperative context. Stage 2 combines local and cooperative perceptions,
selects Fast or Slow reasoning, and generates the signal action
(Algorithm~\ref{alg:vlalight_inference}).

\vspace{-0.8em}
\begin{algorithm}[H]
\caption{Two-Stage Cooperative VLA Inference}
\label{alg:vlalight_inference}
\begin{algorithmic}[1]
\REQUIRE Four-direction videos $\mathbf{V}_i^t$, topology-routed coordination evidence $\mathbf{N}_i^t$, current phase and history, policy $\pi_\theta$
\ENSURE Signal phase $a_i^t$
\STATE Generate the structured local perception $P_i^t$ from the four-direction videos $\mathbf{V}_i^t$
\STATE Route relevant perceptions and coordination evidence through the directed road topology to obtain cooperative context $C_i^t$
\STATE Combine $P_i^t$ and $C_i^t$ into the Stage 2 decision context
\STATE Select $m_i^t\in\{\textnormal{Fast},\textnormal{Slow}\}$ adaptively from the observed traffic context
\STATE Generate the signal decision conditioned on the combined context and selected mode
\STATE Parse the selected signal phase as $a_i^t$
\RETURN $a_i^t$
\end{algorithmic}
\end{algorithm}
\vspace{-0.8em}

\subsection{Construction of Cooperative Information}
\label{app:cooperative_information}

\begin{figure}[H]
    \centering
    \includegraphics[width=0.90\linewidth]{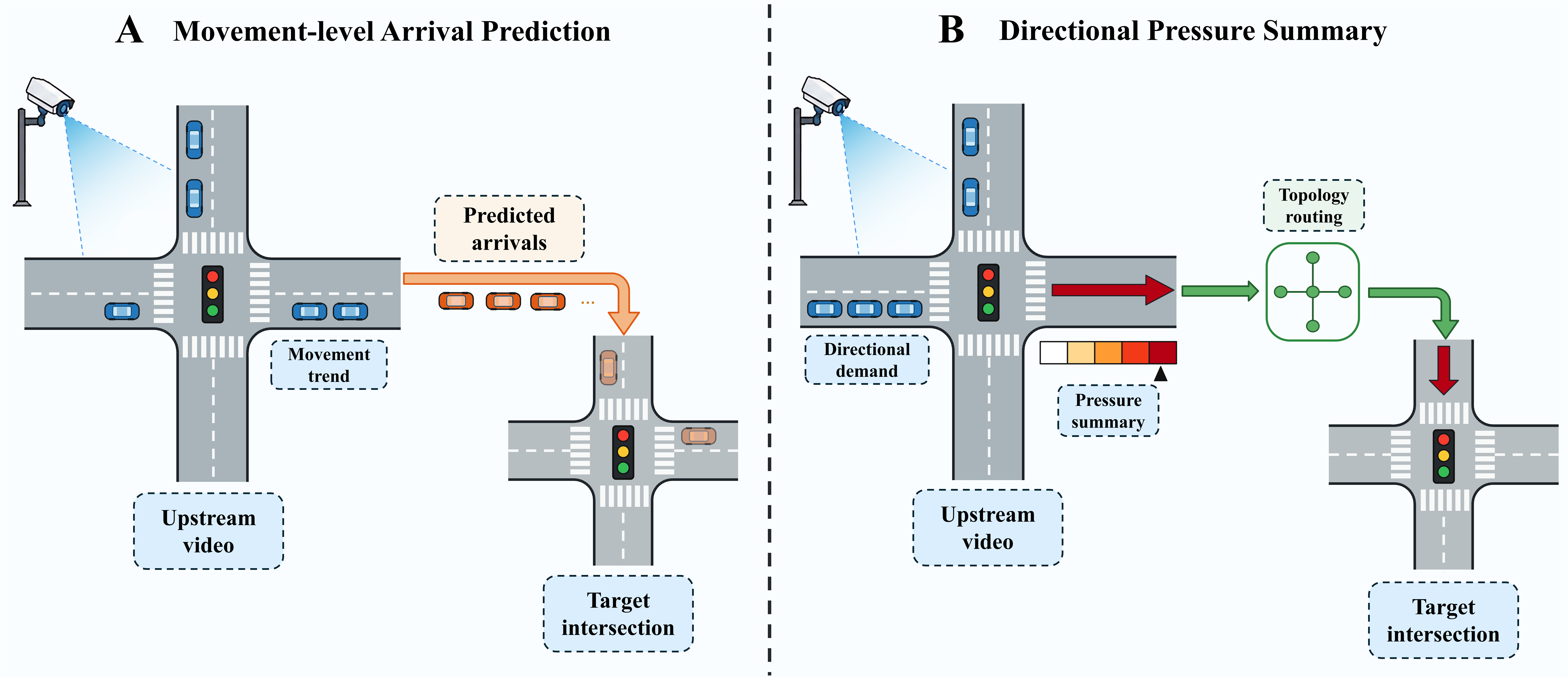}
    \caption{Topology-routed cooperative information: (A) downstream-linked frames estimate potential arrivals; (B) upstream vehicle and queue counts with link travel time summarize directional demand.}
    \label{fig:cooperative_context}
\end{figure}

At each decision step, VLALight routes two complementary signals over the directed
road network (Figure~\ref{fig:cooperative_context}). For each candidate phase, it
follows outgoing lanes to the connected downstream intersection and extracts one
frame from its video window. This frame estimates potential arrivals on the
corresponding target movement, forming the \texttt{local\_coordination} field and
representing vehicles not yet visible locally.

For the \texttt{neighbors} field, Stage 1 perceptions from actual neighboring
intersections are mapped through the movement-route table to the target's approach
directions. Vehicle and queue counts on connected upstream movements are aggregated
by direction and paired with estimated link travel time. This summarizes upstream
pressure, not guaranteed arrivals, because upstream signals control whether and when
vehicles are released. Stage 2 receives both fields alongside the target's local
perception.

\clearpage
\subsection{Case Study}
\label{app:case_study}
\input{figures/case_study/case_study.tex}

\subsection{Prompt Templates}
\label{app:prompt_templates}
\input{figures/prompt_templates.tex}

\paragraph{Code Availability.} The source code is publicly available at \url{https://github.com/usail-hkust/VLALight.git}.

\end{document}

%% file: math_commands.tex
\usepackage{amsmath,amsfonts,bm}

\def\eqref#1{equation~\ref{#1}}

\def\1{\bm{1}}

\DeclareMathAlphabet{\mathsfit}{\encodingdefault}{\sfdefault}{m}{sl}
\SetMathAlphabet{\mathsfit}{bold}{\encodingdefault}{\sfdefault}{bx}{n}



%% file: figures/traffic_pressure_mode_cases.tex

\definecolor{pressureFastDark}{HTML}{2166A5}
\definecolor{pressureFastBg}{HTML}{EEF6FC}
\definecolor{pressureSlowDark}{HTML}{C55A27}
\definecolor{pressureSlowBg}{HTML}{FFF3EC}
\definecolor{pressureInk}{HTML}{263238}
\definecolor{pressureMuted}{HTML}{68737D}
\definecolor{pressureRule}{HTML}{CBD5DE}

\newcommand{\PressureDecisionCue}[2]{%
  \vspace{1.0pt}
  {\fontsize{5.0}{5.75}\selectfont
   \textcolor{#1}{\bfseries Why:} #2\par}}

\newcommand{\PressureKeyReason}[2]{%
  \vspace{1.0pt}
  {\fontsize{5.0}{5.7}\selectfont
   \textcolor{#1}{\bfseries Key reasoning:}\ \textit{#2}\par}}

\newcommand{\PressurePhaseLine}[5]{%
  \noindent\makebox[\linewidth][l]{#1:\quad $v$=#2,\ $q$=#3,\ $\Delta v$=#4,\ $\Delta q$=#5}\par}

\newcommand{\PressureCoopBlock}[3]{%
  \textcolor{#1}{\bfseries Neighbor movement cues:}\par
  #2\par\vspace{0.35pt}
  \textcolor{#1}{\bfseries Neighbor directional pressure:}\par
  #3}

\newcommand{\PressureCasePanel}[9]{%
  {\setlength{\fboxsep}{3.2pt}%
   \fcolorbox{pressureRule}{white}{%
     \begin{minipage}[t]{0.458\textwidth}
       \raggedright
       \color{pressureInk}
       \fontsize{5.25}{6.05}\selectfont
       \colorbox{#2}{%
         \parbox{\dimexpr\linewidth-2\fboxsep\relax}{%
           \textcolor{#1}{\fontsize{7.0}{7.8}\selectfont\bfseries #3}%
           \hfill
           {\textcolor{pressureMuted}{\fontsize{4.65}{5.25}\selectfont #4}}}}
       \vspace{1.8pt}
       \begin{tabular}{@{}p{0.525\linewidth}@{\hspace{1.7mm}}p{0.435\linewidth}@{}}
         \begin{minipage}[t]{\linewidth}
           \textcolor{#1}{\bfseries Local perception}\par\vspace{0.45pt}
           {\fontsize{5.0}{5.6}\selectfont #5}
         \end{minipage}
         &
         \begin{minipage}[t]{\linewidth}
           \textcolor{#1}{\bfseries Cooperative context}\par\vspace{0.45pt}
           {\fontsize{4.8}{5.25}\selectfont #6}
         \end{minipage}
       \end{tabular}\par
       \vspace{1.2pt}
       \colorbox{#2}{%
         \parbox{\dimexpr\linewidth-2\fboxsep\relax}{%
           \fontsize{5.05}{5.8}\selectfont
           \textcolor{#1}{\bfseries #7}\hfill
           \textcolor{pressureMuted}{\bfseries signal}\ \textcolor{#1}{\bfseries #8}}}\par
       #9
     \end{minipage}%
   }}%
}

\begin{figure*}[t]
  \centering
  \begin{tabular}{@{}c@{\hspace{2.5mm}}c@{}}
    \PressureCasePanel{pressureFastDark}{pressureFastBg}
      {Fast 1 $\cdot$ clear local winner}
      {intersection 3--1, step 99}
      {\PressurePhaseLine{ETWT}{19}{19}{0}{0}
       \PressurePhaseLine{\textcolor{pressureFastDark}{\textbf{NTST}}}{\textbf{39}}{\textbf{39}}{\textbf{0}}{\textbf{0}}
       \PressurePhaseLine{ELWL}{8}{8}{0}{0}
       \PressurePhaseLine{NLSL}{0}{0}{0}{0}}
      {\PressureCoopBlock{pressureFastDark}
        {ETWT: 19 vehicles $\cdot$ NTST: 19 vehicles}
        {N: 20 queued $\cdot$ E: 19 queued $\cdot$ ETA: 27 s}}
      {FAST}{NTST}
      {\PressureDecisionCue{pressureFastDark}{The local queue maximum is unambiguous.}}
    &
    \PressureCasePanel{pressureFastDark}{pressureFastBg}
      {Fast 2 $\cdot$ clear local winner}
      {intersection 5--1, step 88}
      {\PressurePhaseLine{ETWT}{14}{0}{$-6$}{$-20$}
       \PressurePhaseLine{\textcolor{pressureFastDark}{\textbf{NTST}}}{\textbf{21}}{\textbf{21}}{\textbf{0}}{\textbf{0}}
       \PressurePhaseLine{ELWL}{1}{1}{0}{0}
       \PressurePhaseLine{NLSL}{0}{0}{0}{0}}
      {\PressureCoopBlock{pressureFastDark}
        {ETWT: 1 vehicle}
        {E: 1 queued $\cdot$ ETA: 27 s}}
      {FAST}{NTST}
      {\PressureDecisionCue{pressureFastDark}{One phase dominates visible and stopped vehicles.}}
    \\
    [-0.2mm]
    \PressureCasePanel{pressureSlowDark}{pressureSlowBg}
      {Slow 1 $\cdot$ competing local queues}
      {intersection 2--1, step 40}
      {\PressurePhaseLine{\textcolor{pressureSlowDark}{\textbf{ETWT}}}{\textbf{20}}{\textbf{19}}{\textbf{$+4$}}{\textbf{$+10$}}
       \PressurePhaseLine{NTST}{21}{19}{$-1$}{$-1$}
       \PressurePhaseLine{ELWL}{5}{5}{0}{0}
       \PressurePhaseLine{NLSL}{0}{0}{0}{0}}
      {\PressureCoopBlock{pressureSlowDark}
        {ETWT: 11 vehicles $\cdot$ NTST: 1 vehicle}
        {E: 20 queued $\cdot$ N: 12 queued $\cdot$ ETA: 27 s}}
      {SLOW}{ETWT}
      {\PressureKeyReason{pressureSlowDark}{Given ETWT has the largest visible demand, worsening trend, and strong local coordination plus east neighbor pressure, it is clearly the most effective choice.}}
    &
    \PressureCasePanel{pressureSlowDark}{pressureSlowBg}
      {Slow 2 $\cdot$ close local competition}
      {intersection 7--10, step 62}
      {\PressurePhaseLine{ETWT}{4}{3}{$+1$}{$+1$}
       \PressurePhaseLine{NTST}{0}{0}{$-5$}{$-2$}
       \PressurePhaseLine{\textcolor{pressureSlowDark}{\textbf{ELWL}}}{\textbf{5}}{\textbf{3}}{\textbf{$+2$}}{\textbf{$+1$}}
       \PressurePhaseLine{NLSL}{2}{2}{0}{$+1$}}
      {\PressureCoopBlock{pressureSlowDark}
        {ETWT: 1 vehicle $\cdot$ ELWL: 1 vehicle}
        {N: 2 queued $\cdot$ S: 2 queued $\cdot$ ETA: 27 s}}
      {SLOW}{ELWL}
      {\PressureKeyReason{pressureSlowDark}{Given ELWL’s high demand, queue, positive trend, and long waiting time, it is the most effective choice.}}
  \end{tabular}
  \caption{Case study on adapative reasoning.}
  \label{fig:pressure_mode_cases}
\vspace{-15pt}
\end{figure*}

%% file: figures/case_study/case_study.tex

\graphicspath{{figures/case_study/images/}}

\definecolor{caseSlate}{HTML}{34495E}
\definecolor{caseNavy}{HTML}{0E2841}
\definecolor{caseOuterGray}{HTML}{ECF0F1}
\definecolor{casePerceptionFill}{HTML}{F1F4D6}
\definecolor{caseFastFill}{HTML}{FDEBD0}
\definecolor{caseSlowFill}{HTML}{CADEEC}
\definecolor{caseMuted}{HTML}{5F6B73}
\definecolor{caseSelected}{HTML}{0072B2}
\definecolor{caseLabelBlue}{HTML}{4E6E83}
\definecolor{casePerceptionBlockFill}{HTML}{FAFBEF}
\definecolor{casePerceptionBlockLine}{HTML}{D7DEC9}
\definecolor{caseReasonBlockFill}{HTML}{EAF3F8}
\definecolor{caseReasonBlockLine}{HTML}{B9CEDB}

\newcommand{\CaseStudyFigFont}{\fontsize{7}{8.2}\selectfont}
\newcommand{\CaseStudyFigTitle}{\fontsize{8.4}{9.2}\selectfont\bfseries}
\newcommand{\CaseStudyCornerRadius}{3pt}
\newcommand{\CaseStudySlowSelectedPhaseGap}{0pt}
\newcommand{\CaseStudyObservationTrim}{36bp}
\newcommand{\CaseStudyObservationTopSpace}{1.5pt}
\newcommand{\CaseStudyObservationRowSpace}{0.8mm}

\newcommand{\CaseStudyFrame}[4]{%
  \begin{tikzpicture}[baseline=(frame.south)]
    \node[anchor=north west,inner sep=0,draw=caseSlate,line width=.3pt] (north) at (0,0)
      {\includegraphics[width=.81cm,trim=0 0 0 \CaseStudyObservationTrim,clip]{figures/case_study/images/#1_N_#2}};
    \node[anchor=north west,inner sep=0,draw=caseSlate,line width=.3pt] (east) at (north.north east)
      {\includegraphics[width=.81cm,trim=0 0 0 \CaseStudyObservationTrim,clip]{figures/case_study/images/#1_E_#2}};
    \node[anchor=north west,inner sep=0,draw=caseSlate,line width=.3pt] (west) at (north.south west)
      {\includegraphics[width=.81cm,trim=0 0 0 \CaseStudyObservationTrim,clip]{figures/case_study/images/#1_W_#2}};
    \node[anchor=north west,inner sep=0,draw=caseSlate,line width=.3pt] (south) at (east.south west)
      {\includegraphics[width=.81cm,trim=0 0 0 \CaseStudyObservationTrim,clip]{figures/case_study/images/#1_S_#2}};
    \node[fit=(north)(east)(west)(south),inner sep=0,draw=caseSlate,line width=.65pt] (frame) {};
    \foreach \tile/\labeltext in {north/N,east/E,west/W,south/S}{
      \node[anchor=north west,fill=caseLabelBlue,rounded corners=1.2pt,text=white,
            inner sep=1pt,font=\fontsize{7}{7}\selectfont\bfseries]
        at ([xshift=.4pt,yshift=-.4pt]\tile.north west) {\labeltext};
    }
    \node[anchor=south,fill=caseLabelBlue,rounded corners=1.2pt,text=white,
          minimum width=1.63cm,inner ysep=1.15pt,
          font=\fontsize{7}{7}\selectfont\bfseries]
      at ([yshift=.3pt]frame.north) {F#3\enspace $t=#4$ s};
  \end{tikzpicture}%
}

\newcommand{\CaseStudyDecisionField}[2]{%
  \noindent #1:\enspace\textbf{#2}\par
}

\newcommand{\CaseStudyInnerBlock}[3]{%
  \noindent\hspace*{.18cm}%
  \begin{tikzpicture}[baseline=(inner.base)]
    \node[anchor=base west,rounded corners=\CaseStudyCornerRadius,draw=#1,fill=#2,
          line width=.3pt,inner sep=2pt,minimum height=0pt,text width=6.65cm,
          align=left,font=\CaseStudyFigFont] (inner) {#3};
  \end{tikzpicture}%
  \par\vspace{1.5pt}%
}

\tikzset{
  caseStudy/perception/.style={
    anchor=north west,draw=black,rounded corners=\CaseStudyCornerRadius,
    fill=casePerceptionFill,line width=.65pt,inner sep=4pt,text width=7.10cm,
    minimum height=4.10cm,align=left,font=\CaseStudyFigFont
  },
  caseStudy/decision/.style={
    anchor=north west,draw=black,rounded corners=\CaseStudyCornerRadius,
    line width=.65pt,inner sep=4pt,text width=7.10cm,minimum height=2.85cm,
    align=left,font=\CaseStudyFigFont
  },
  caseStudy/process/.style={
    draw=caseSlate,dashed,rounded corners=\CaseStudyCornerRadius,
    fill=caseOuterGray,line width=.5pt,inner sep=3pt
  },
  caseStudy/observation/.style={
    anchor=north west,draw=caseSlate,rounded corners=\CaseStudyCornerRadius,
    fill=white,line width=.7pt,inner sep=3pt
  }
}

\newcommand{\VLALightFastCaseGraphic}{%
\begin{tikzpicture}
  \node[anchor=south west,text=caseNavy,font=\fontsize{9}{10}\selectfont\bfseries]
    at (0,.13) {Fast case: a clear local-demand winner};
  \node[anchor=south east,text=caseMuted,font=\CaseStudyFigFont]
    at (13.90,.13) {New York $\cdot$ intersection 1-19 $\cdot$ step 54};

  \node[caseStudy/observation] (fastobs) at (0,0) {\begin{minipage}{5.20cm}
    {\CaseStudyFigTitle Video observations}\hfill{\CaseStudyFigFont 6 frames, 1 s apart}\par
    \vspace{\CaseStudyObservationTopSpace}
    \setlength{\tabcolsep}{.40mm}
    \begin{tabular}{@{}ccc@{}}
      \CaseStudyFrame{fast}{frame_000000_t00000.000s.jpg}{1}{0} &
      \CaseStudyFrame{fast}{frame_000001_t00001.000s.jpg}{2}{1} &
      \CaseStudyFrame{fast}{frame_000002_t00002.000s.jpg}{3}{2} \\[\CaseStudyObservationRowSpace]
      \CaseStudyFrame{fast}{frame_000003_t00003.000s.jpg}{4}{3} &
      \CaseStudyFrame{fast}{frame_000004_t00004.000s.jpg}{5}{4} &
      \CaseStudyFrame{fast}{frame_000005_t00005.000s.jpg}{6}{5}
    \end{tabular}
  \end{minipage}};

  \node[caseStudy/perception] (fastperc) at (6.23,0) {\begin{minipage}{7.10cm}
    {\CaseStudyFigTitle Perception}\par\vspace{1.5pt}
    Current phase: \textbf{ETWT}\par\vspace{1.5pt}
    \textbf{Local perception}\par\vspace{1pt}
    \CaseStudyInnerBlock{casePerceptionBlockLine}{casePerceptionBlockFill}{%
      \raggedright
      \renewcommand{\arraystretch}{1.08}
      \begin{tabular*}{6.55cm}{@{\extracolsep{\fill}}lccccc@{\hspace{4pt}}}
        Phase & $v$ & $q$ & $\Delta v$ & $\Delta q$ & age \\
        ETWT & 0 & 0 & $-6$ & $-6$ & 0 \\
        \textcolor{caseSelected}{\textbf{NTST}} & \textbf{9} & \textbf{9} &
          \textbf{$+4$} & \textbf{$+8$} & \textbf{2} \\
        ELWL & 4 & 3 & $+2$ & $+2$ & 1 \\
        NLSL & 0 & 0 & 0 & 0 & 0
      \end{tabular*}%
    }
    \textbf{Cooperative perception}\par\vspace{1pt}
    \CaseStudyInnerBlock{casePerceptionBlockLine}{casePerceptionBlockFill}{%
      \raggedright
      \textbf{1. Direct:} arrivals are $0$ for all phases.\par
      \textbf{2. Neighbor:} S $4/2$, N $0/0$, E $2/1$ $(v/q)$; ETA $27$ s.
    }
  \end{minipage}};

  \node[caseStudy/decision,fill=caseFastFill,below=1.3mm of fastperc.south west,
        anchor=north west] (fastdec) {\begin{minipage}[t][2.50cm][t]{7.10cm}
      {\CaseStudyFigTitle Decision}\par\vspace*{\fill}
      \CaseStudyDecisionField{Routing model}{FAST}
      \vspace{7pt}
      \CaseStudyDecisionField{Selected phase}{NTST}
      \vspace*{\fill}
    \end{minipage}};
  \begin{scope}[on background layer]
    \node[caseStudy/process,fit=(fastperc)(fastdec)] (fastprocess) {};
  \end{scope}
  \draw[-{Latex[length=2.3mm,width=1.7mm]},line width=1.15pt,caseNavy]
    ([xshift=1mm]fastobs.east) -- ([xshift=-1mm]fastprocess.west |- fastobs.east);
\end{tikzpicture}%
}

\newcommand{\VLALightSlowCaseGraphic}{%
\begin{tikzpicture}
  \node[anchor=south west,text=caseNavy,font=\fontsize{9}{10}\selectfont\bfseries]
    at (0,.13) {Slow case: coordination resolves competing queues};
  \node[anchor=south east,text=caseMuted,font=\CaseStudyFigFont]
    at (13.90,.13) {New York $\cdot$ intersection 3-1 $\cdot$ step 52};

  \node[caseStudy/observation] (slowobs) at (0,0) {\begin{minipage}{5.20cm}
    {\CaseStudyFigTitle Video observations}\hfill{\CaseStudyFigFont 6 frames, 1 s apart}\par
    \vspace{\CaseStudyObservationTopSpace}
    \setlength{\tabcolsep}{.40mm}
    \begin{tabular}{@{}ccc@{}}
      \CaseStudyFrame{slow}{frame_000000_t00000.000s.jpg}{1}{0} &
      \CaseStudyFrame{slow}{frame_000001_t00001.000s.jpg}{2}{1} &
      \CaseStudyFrame{slow}{frame_000002_t00002.000s.jpg}{3}{2} \\[\CaseStudyObservationRowSpace]
      \CaseStudyFrame{slow}{frame_000003_t00003.000s.jpg}{4}{3} &
      \CaseStudyFrame{slow}{frame_000004_t00004.000s.jpg}{5}{4} &
      \CaseStudyFrame{slow}{frame_000005_t00005.000s.jpg}{6}{5}
    \end{tabular}
  \end{minipage}};

  \node[caseStudy/perception] (slowperc) at (6.23,0) {\begin{minipage}{7.10cm}
    {\CaseStudyFigTitle Perception}\par\vspace{1.5pt}
    Current phase: \textbf{NTST}\par\vspace{1.5pt}
    \textbf{Local perception}\par\vspace{1pt}
    \CaseStudyInnerBlock{casePerceptionBlockLine}{casePerceptionBlockFill}{%
      \raggedright
      \renewcommand{\arraystretch}{1.08}
      \begin{tabular*}{6.55cm}{@{\extracolsep{\fill}}lccccc@{\hspace{4pt}}}
        Phase & $v$ & $q$ & $\Delta v$ & $\Delta q$ & age \\
        \textcolor{caseSelected}{\textbf{ETWT}} & \textbf{20} & \textbf{20} &
          \textbf{0} & \textbf{0} & \textbf{2} \\
        NTST & 19 & 19 & 0 & 0 & 0 \\
        ELWL & 1 & 1 & 0 & 0 & 0 \\
        NLSL & 19 & 19 & 0 & 0 & 0
      \end{tabular*}%
    }
    \textbf{Cooperative perception}\par\vspace{1pt}
    \CaseStudyInnerBlock{casePerceptionBlockLine}{casePerceptionBlockFill}{%
      \raggedright
      \textbf{1. Direct:} ETWT $19$; all other phases $0$.\par
      \textbf{2. Neighbor:} W $0/0$, N $8/0$, E $10/8$ $(v/q)$; ETA $27$ s.
    }
  \end{minipage}};

  \node[caseStudy/decision,fill=caseSlowFill,below=1.3mm of slowperc.south west,
        anchor=north west] (slowdec) {\begin{minipage}[t][2.50cm][t]{7.10cm}
      {\CaseStudyFigTitle Decision}\par\vspace{3pt}
      \CaseStudyDecisionField{Routing model}{SLOW}
      \vspace{2pt}
      \noindent Reasoning:\par\vspace{1pt}
      \CaseStudyInnerBlock{caseReasonBlockLine}{caseReasonBlockFill}{%
        \noindent\hspace*{.08cm}\parbox[t]{6.38cm}{\raggedright
          Local demand is nearly tied (ETWT $20$ vs. NTST/NLSL $19$).
          NTST has just been served; ETWT has waited two intervals. Its $19$
          coordinated arrivals and east-side queue break the tie toward east-west service.}%
      }
      \vspace{\CaseStudySlowSelectedPhaseGap}
      \CaseStudyDecisionField{Selected phase}{ETWT}
    \end{minipage}};
  \begin{scope}[on background layer]
    \node[caseStudy/process,fit=(slowperc)(slowdec)] (slowprocess) {};
  \end{scope}
  \draw[-{Latex[length=2.3mm,width=1.7mm]},line width=1.15pt,caseNavy]
    ([xshift=1mm]slowobs.east) -- ([xshift=-1mm]slowprocess.west |- slowobs.east);
\end{tikzpicture}%
}

We present two decision steps from the New York network to illustrate how VLALight adapts
reasoning depth to traffic complexity. Each case contains six one-second-spaced
observations, and each observation combines the north, east, west, and south views of the
target intersection. The figures expose the video-grounded local perception, routed
cooperative perception, routing mode, and selected phase; the following paragraphs explain
why the two cases take different decision paths.

\begin{figure}[H]
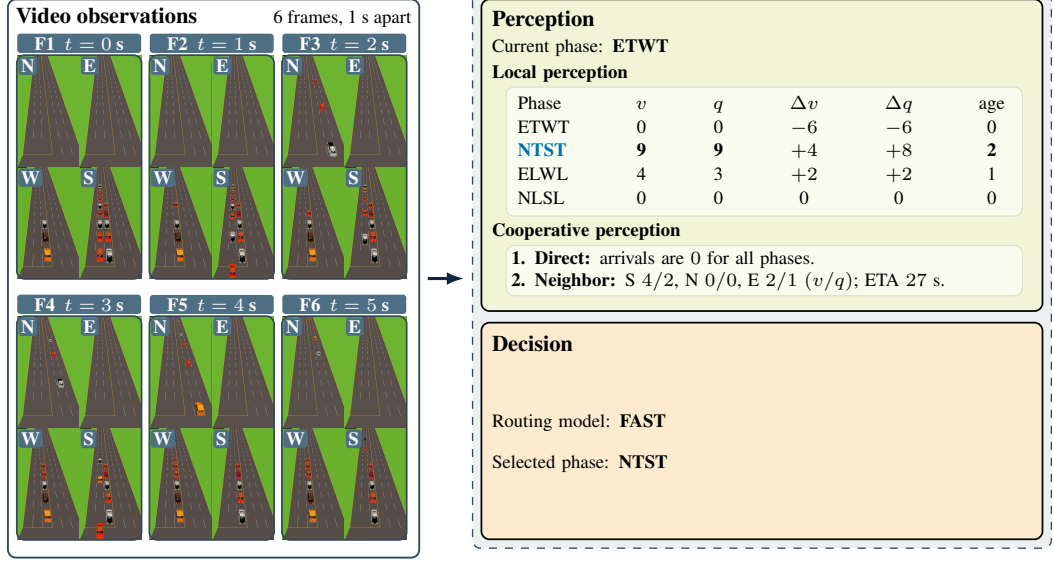

  \centering
  \resizebox{\linewidth}{!}{\VLALightFastCaseGraphic}
  \caption{\textbf{Fast case.} A clear local-demand winner leads to fast routing.}
  \label{fig:case-fast}
\end{figure}

\begin{figure}[H]
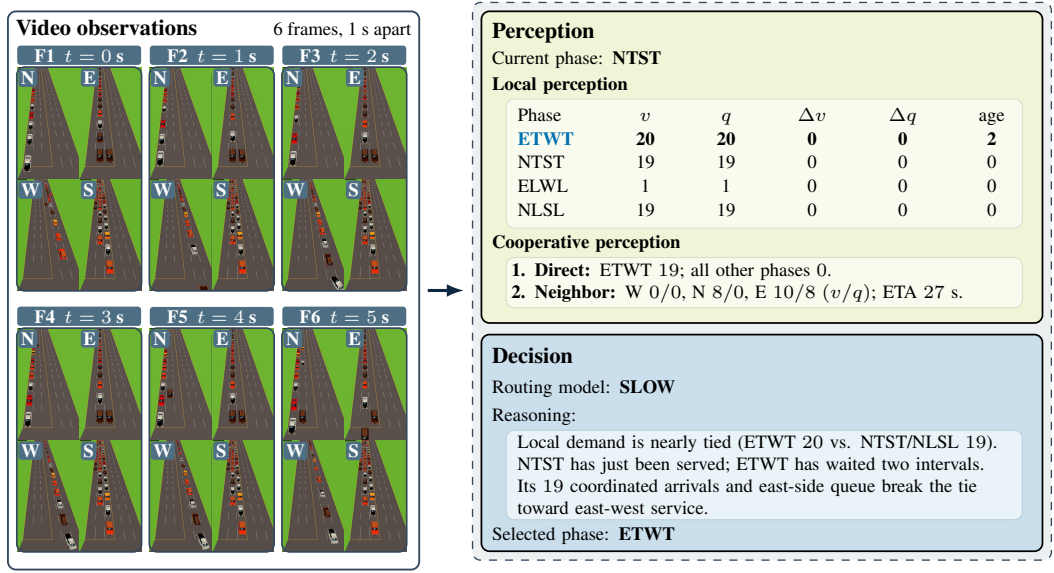

  \centering
  \resizebox{\linewidth}{!}{\VLALightSlowCaseGraphic}
  \caption{\textbf{Slow case.} Competing local and cooperative evidence triggers slow reasoning.}
  \label{fig:case-slow}
\end{figure}

\subsubsection{Fast Case}
At intersection 1-19 (step 54), NTST is the clear local
winner: it has the largest visible and queued demand ($v/q=9/9$), the strongest recent
increase ($\Delta v/\Delta q=+4/+8$), and the longest unserved age. The upstream context
contains no direct phase-level arrivals and only modest neighbour pressure. VLALight can
therefore trust the video-grounded local evidence, route the decision to the fast branch,
and select NTST without spending tokens on deliberation. This case shows how the policy
preserves low-latency control when one phase is already supported by consistent evidence.

\subsubsection{Slow Case}
At intersection 3-1 (step 52), local demand is ambiguous:
ETWT has $v/q=20/20$ and has waited for two intervals, while NTST and NLSL each have
$v/q=19/19$ and were served more recently. The routed context adds 19 direct ETWT
arrivals and upstream pressure from the east, north, and west approaches. VLALight invokes
the slow branch to reconcile these local and cooperative signals, identifies the waiting
and arrival advantage of ETWT, and selects ETWT. This case demonstrates that the adaptive
policy reserves explicit reasoning for decisions where network context changes the local
preference rather than applying deliberation uniformly.

%% file: figures/prompt_templates.tex
\makeatletter
\def\verbatim@font{\ttfamily\fontsize{6.2}{7.2}\selectfont}
\makeatother

This appendix records the prompt templates used by the video-grounded controller. The
templates are shown in normalized form: runtime-specific values such as the intersection
identifier, video objects, coordination frames, perception JSON, and neighbour summaries
are inserted into the indicated placeholders. The output blocks are kept explicit because
they define the interface between visual perception and signal control.

\subsubsection{Stage 1: Video-Grounded Perception}

The perception prompt supplies four direction-specific videos and the available upstream
coordination frames. It asks the model to return traffic features for all four candidate
phases and to keep local observations separate from predicted arrivals.

\begin{verbatim}
[system]
You are a visual traffic perception model for a four-way intersection.

You receive four direction-specific videos of the current intersection and any
available upstream coordination frames. Infer only the structured traffic features
supported by the visual evidence.

Right-turn movements are permanently permissive and are not controlled by the
four candidate signal phases.

Return exactly one <perception>...</perception> block. Do not output a mode,
reasoning, signal decision, or any text outside the perception block.

[user]
Intersection: <INTERSECTION_ID>
Current phase: <CURRENT_PHASE>

Visual inputs:
- East approach video: <VIDEO_E>
- West approach video: <VIDEO_W>
- North approach video: <VIDEO_N>
- South approach video: <VIDEO_S>

Upstream coordination frames, keyed by target entry direction:
<COORDINATION_FRAMES>

Use every supplied coordination frame. The frame label is the target
intersection's entry direction. Attribute counts as follows:
E -> ET and EL; W -> WT and WL; N -> NT and NL; S -> ST and SL.
Missing directions indicate unavailable upstream visual evidence, usually at a
road-network boundary. Do not treat a missing frame as zero vehicles.
Keep coordinated-arrival counts separate from current_v and current_q.

Controller-provided persistent-demand history:
<UNSERVED_HISTORY>

Temporal convention:
- Each approach video contains six synchronized frames sampled every 5 seconds:
  t=5, 10, 15, 20, 25, and 30 seconds in the current decision window.
- The decision is made at t=30 seconds.
- current_v and current_q describe the latest observation at t=30 seconds.
- dv and dq compare the latest and earliest observations.
- For a boundary movement, the coordinated-arrival estimate may use the
  t=30 versus t=15 observation; do not interpret it as a general demand trend.

For every candidate phase, estimate:
- the total visible vehicles and the two movement-level counts;
- the stopped subset of those vehicles;
- the recent changes in visible demand and queue size;
- the potential arrivals from available upstream frames.

Current V is the total visible service demand in the two movements released by a
phase. Its movement breakdown must not be added again. Current q is a subset of
current V. Coordinated arrivals are future-pressure estimates and must not be
added to local visible demand.

Output exactly one structured perception block and nothing else:

<perception>
{
  "current_phase": "<CURRENT_PHASE>",
  "phases": {
    "ETWT": {
      "v": [<ET_V>, <WT_V>], "q": [<ET_Q>, <WT_Q>],
      "dv": <ETWT_DV>, "dq": <ETWT_DQ>, "age": <ETWT_AGE>,
      "coord": {
        "ET": {"count": <ET_ARRIVALS>, "is_boundary": "<yes/no>"},
        "WT": {"count": <WT_ARRIVALS>, "is_boundary": "<yes/no>"}
      }
    },
    "NTST": {
      "v": [<NT_V>, <ST_V>], "q": [<NT_Q>, <ST_Q>],
      "dv": <NTST_DV>, "dq": <NTST_DQ>, "age": <NTST_AGE>,
      "coord": {
        "NT": {"count": <NT_ARRIVALS>, "is_boundary": "<yes/no>"},
        "ST": {"count": <ST_ARRIVALS>, "is_boundary": "<yes/no>"}
      }
    },
    "ELWL": {
      "v": [<EL_V>, <WL_V>], "q": [<EL_Q>, <WL_Q>],
      "dv": <ELWL_DV>, "dq": <ELWL_DQ>, "age": <ELWL_AGE>,
      "coord": {
        "EL": {"count": <EL_ARRIVALS>, "is_boundary": "<yes/no>"},
        "WL": {"count": <WL_ARRIVALS>, "is_boundary": "<yes/no>"}
      }
    },
    "NLSL": {
      "v": [<NL_V>, <SL_V>], "q": [<NL_Q>, <SL_Q>],
      "dv": <NLSL_DV>, "dq": <NLSL_DQ>, "age": <NLSL_AGE>,
      "coord": {
        "NL": {"count": <NL_ARRIVALS>, "is_boundary": "<yes/no>"},
        "SL": {"count": <SL_ARRIVALS>, "is_boundary": "<yes/no>"}
      }
    }
  }
}
</perception>
\end{verbatim}

Here $v$ denotes visible vehicle count, $q$ denotes stopped-vehicle count, $dv$
and $dq$ denote recent changes, and $age$ records persistent unmet demand. The
movement order is fixed as ETWT=[ET,WT], NTST=[NT,ST], ELWL=[EL,WL], and
NLSL=[NL,SL].

\subsubsection{Stage 2: Routed Signal Decision}

The decision prompt receives the structured local perception and the routed
cooperative context. It constrains the output to a valid phase and selects Fast
mode when one phase is clearly preferred, while using Slow mode when local and
cooperative evidence require an explicit comparison.

\begin{verbatim}
[system]
You are the decision stage of a traffic signal controller.
The perception JSON below was produced by Stage 1 and has already been routed
across the intersection. Choose exactly one phase from ETWT, NTST, ELWL, NLSL.

Return either:
<mode>fast</mode>
<signal>PHASE</signal>

or:
<mode>slow</mode>
<reasoning>brief comparison</reasoning>
<signal>PHASE</signal>

Do not emit markdown or text outside the required tags.

[user]
You are an intelligent traffic signal controller for a four-phase intersection.
Use the local and cooperative perceptions below to decide which candidate phase
should be served next.

The controller operates in 30-second cycles: 25 seconds of green followed by a
5-second signal-transition interval. The decision is made immediately before
the transition, and the selected phase receives the next green interval.

<local_perception>
{
  "current_phase": "<CURRENT_PHASE>",
  "phases": {
    "ETWT": {"current_v": <ETWT_V>, "current_q": <ETWT_Q>,
              "dv": <ETWT_DV>, "dq": <ETWT_DQ>, "unserved_age": <ETWT_AGE>},
    "NTST": {"current_v": <NTST_V>, "current_q": <NTST_Q>,
              "dv": <NTST_DV>, "dq": <NTST_DQ>, "unserved_age": <NTST_AGE>},
    "ELWL": {"current_v": <ELWL_V>, "current_q": <ELWL_Q>,
              "dv": <ELWL_DV>, "dq": <ELWL_DQ>, "unserved_age": <ELWL_AGE>},
    "NLSL": {"current_v": <NLSL_V>, "current_q": <NLSL_Q>,
              "dv": <NLSL_DV>, "dq": <NLSL_DQ>, "unserved_age": <NLSL_AGE>}
  }
}
</local_perception>

<cooperative_perception>
{
  "local_coordination": {
    "ETWT": <ETWT_POTENTIAL_ARRIVALS>, "NTST": <NTST_POTENTIAL_ARRIVALS>,
    "ELWL": <ELWL_POTENTIAL_ARRIVALS>, "NLSL": <NLSL_POTENTIAL_ARRIVALS>
  },
  "neighbors": {
    "<DIRECTION>": {"total_v": <NEIGHBOR_V>, "total_q": <NEIGHBOR_Q>,
                      "travel_time_s": <TRAVEL_TIME>}
  }
}
</cooperative_perception>

Local-field meanings:
- current_phase is the phase currently receiving service.
- current_v is the visible serviceable vehicle count for a phase.
- current_q is the stopped-vehicle subset of current_v.
- dv and dq are recent changes in visible demand and queue size.
- unserved_age measures the number of decision intervals since service.

Cooperative-field meanings:
- local_coordination contains potential arrivals for each target phase.
  Keep these arrivals separate from current_v and current_q.
- neighbors contains broad directional pressure from connected intersections.
  Do not convert it into a fixed target phase or assume that all vehicles will
  definitely arrive, because the neighbour's future release is unknown.
- North/south neighbour pressure can adjust the priority of NTST and NLSL;
  east/west pressure can adjust the priority of ETWT and ELWL.
- travel_time_s indicates when broad neighbour pressure may reach the target.
- Missing directions are boundary directions and should be ignored.

Decision guidance:
- Prefer phases that can release the largest visible demand and queue pressure.
- Prioritize sustained or worsening pressure over brief fluctuations.
- Use direct local_coordination, persistent unmet demand, and queue urgency to
  break ties between otherwise comparable phases.
- Choose fast mode when one candidate phase is clearly preferable.
- Choose slow mode when multiple phases are comparable or when local trends,
  direct coordination, persistent demand, and neighbour context disagree.

Critical output format:

Fast mode:
<mode>fast</mode>
<signal>SELECTED_PHASE</signal>

Slow mode:
<mode>slow</mode>
<reasoning>concise decision reasoning</reasoning>
<signal>SELECTED_PHASE</signal>

Do not repeat the perception JSON. Do not output text before <mode> or after
</signal>. SELECTED_PHASE must be exactly ETWT, NTST, ELWL, or NLSL.
\end{verbatim}

\paragraph{Perception-to-Decision Interface.}
Stage 1 returns a structured \texttt{<perception>} object containing movement-level
observations for the four candidate phases. Before Stage 2, the router aggregates
movement-level vehicle and queue counts into phase-level \texttt{current\_v} and
\texttt{current\_q}, carries over \texttt{dv} and \texttt{dq}, and maps
\texttt{age} to \texttt{unserved\_age}. The movement-level \texttt{coord} entries
are routed separately as phase-aligned potential arrivals in
\texttt{local\_coordination}. The resulting local summary and routed neighbor
summaries are provided through two parallel blocks, \texttt{<local\_perception>}
and \texttt{<cooperative\_perception>}, respectively. Stage 2 combines these
structured inputs to select the reasoning mode and signal phase, and outputs the
corresponding mode, optional reasoning, and signal tags.

%% file: iclr2027_conference.bbl
\begin{thebibliography}{46}
\providecommand{\natexlab}[1]{#1}
\providecommand{\url}[1]{\texttt{#1}}
\expandafter\ifx\csname urlstyle\endcsname\relax
  \providecommand{\doi}[1]{doi: #1}\else
  \providecommand{\doi}{doi: \begingroup \urlstyle{rm}\Url}\fi

\bibitem[Abdulhai et~al.(2003)Abdulhai, Pringle, and Karakoulas]{abdulhai2003rl}
Baher Abdulhai, Rob Pringle, and Grigoris~J. Karakoulas.
\newblock Reinforcement learning for true adaptive traffic signal control.
\newblock \emph{Journal of Transportation Engineering}, 129\penalty0 (3):\penalty0 278--285, 2003.
\newblock \doi{10.1061/(ASCE)0733-947X(2003)129:3(278)}.

\bibitem[Brohan et~al.(2023)Brohan, Brown, Carbajal, Chebotar, Chen, Choromanski, et~al.]{brohan2023rt2}
Anthony Brohan, Noah Brown, Justice Carbajal, Yevgen Chebotar, Xi~Chen, Krzysztof Choromanski, et~al.
\newblock {RT-2}: Vision-language-action models transfer web knowledge to robotic control.
\newblock \emph{arXiv preprint arXiv:2307.15818}, 2023.

\bibitem[Chen et~al.(2020)Chen, Wei, Xu, Zheng, Yang, Xiong, Xu, and Li]{chen2020mplight}
Chacha Chen, Hua Wei, Nan Xu, Guanjie Zheng, Ming Yang, Yuanhao Xiong, Kai Xu, and Zhenhui Li.
\newblock Toward a thousand lights: Decentralized deep reinforcement learning for large-scale traffic signal control.
\newblock In \emph{Proceedings of the AAAI Conference on Artificial Intelligence}, volume~34, pp.\  3414--3421, 2020.
\newblock \doi{10.1609/aaai.v34i04.5744}.

\bibitem[Devailly et~al.(2022)Devailly, Larocque, and Charlin]{devailly2022igrl}
Fran{\c{c}}ois-Xavier Devailly, Denis Larocque, and Laurent Charlin.
\newblock {IG-RL}: Inductive graph reinforcement learning for massive-scale traffic signal control.
\newblock \emph{IEEE Transactions on Intelligent Transportation Systems}, 23\penalty0 (7):\penalty0 7496--7507, 2022.
\newblock \doi{10.1109/TITS.2021.3070835}.

\bibitem[Hunt et~al.(1982)Hunt, Robertson, Bretherton, and Royle]{hunt1982scoot}
P.~B. Hunt, D.~I. Robertson, R.~D. Bretherton, and M.~C. Royle.
\newblock The {SCOOT} on-line traffic signal optimisation technique.
\newblock \emph{Traffic Engineering \& Control}, 23\penalty0 (4), 1982.

\bibitem[Kim et~al.(2024)Kim, Pertsch, Karamcheti, Xiao, Balakrishna, Nair, Rafailov, Foster, Lam, Sanketi, Vuong, Kollar, Burchfiel, Tedrake, Sadigh, Levine, Liang, and Finn]{kim2024openvla}
Moo~Jin Kim, Karl Pertsch, Siddharth Karamcheti, Ted Xiao, Ashwin Balakrishna, Suraj Nair, Rafael Rafailov, Ethan Foster, Grace Lam, Pannag Sanketi, Quan Vuong, Thomas Kollar, Benjamin Burchfiel, Russ Tedrake, Dorsa Sadigh, Sergey Levine, Percy Liang, and Chelsea Finn.
\newblock {OpenVLA}: An open-source vision-language-action model.
\newblock \emph{arXiv preprint arXiv:2406.09246}, 2024.

\bibitem[Koonce et~al.(2008)Koonce, Rodegerdts, Lee, Quayle, Beaird, Braud, Bonneson, Tarnoff, and Urbanik]{koonce2008timing}
Peter Koonce, Lee Rodegerdts, Kevin Lee, Shaun Quayle, Scott Beaird, Cade Braud, Jim Bonneson, Phil Tarnoff, and Tom Urbanik.
\newblock Traffic signal timing manual.
\newblock Technical Report FHWA-HOP-08-024, Federal Highway Administration, 2008.

\bibitem[Lai et~al.(2025)Lai, Xu, Zhang, Liu, and Xiong]{lai2025llmlight}
Siqi Lai, Zhao Xu, Weijia Zhang, Hao Liu, and Hui Xiong.
\newblock {LLMLight}: Large language models as traffic signal control agents.
\newblock In \emph{Proceedings of the 31st ACM SIGKDD Conference on Knowledge Discovery and Data Mining V.1}, pp.\  2335--2346, 2025.
\newblock \doi{10.1145/3690624.3709379}.

\bibitem[Liang et~al.(2022)Liang, Su, Fang, and Zhong]{liang2022oam}
Enming Liang, Zicheng Su, Chilin Fang, and Renxin Zhong.
\newblock {OAM}: An option-action reinforcement learning framework for universal multi-intersection control.
\newblock In \emph{Proceedings of the AAAI Conference on Artificial Intelligence}, volume~36, pp.\  4550--4558, 2022.
\newblock \doi{10.1609/aaai.v36i4.20378}.

\bibitem[Liu et~al.(2024)Liu, Liu, Wang, An, Li, Zhou, Yang, Zhang, Guo, and Zhang]{liu2024robomamba}
Jiaming Liu, Mengzhen Liu, Zhenyu Wang, Pengju An, Xiaoqi Li, Kaichen Zhou, Senqiao Yang, Renrui Zhang, Yandong Guo, and Shanghang Zhang.
\newblock {RoboMamba}: Efficient vision-language-action model for robotic reasoning and manipulation.
\newblock In \emph{Advances in Neural Information Processing Systems}, volume~37, pp.\  40085--40110, 2024.
\newblock \doi{10.52202/079017-1266}.

\bibitem[Liu et~al.(2026)Liu, Dong, Lu, Diao, Belcak, Liu, Chen, Yin, Wang, Cheng, Choi, Kautz, and Molchanov]{liu2026gdpogrouprewarddecouplednormalization}
Shih-Yang Liu, Xin Dong, Ximing Lu, Shizhe Diao, Peter Belcak, Mingjie Liu, Min-Hung Chen, Hongxu Yin, Yu-Chiang~Frank Wang, Kwang-Ting Cheng, Yejin Choi, Jan Kautz, and Pavlo Molchanov.
\newblock {GDPO}: Group reward-decoupled normalization policy optimization for multi-reward {RL} optimization.
\newblock \emph{arXiv preprint arXiv:2601.05242}, 2026.

\bibitem[Lopez et~al.(2018)Lopez, Behrisch, Bieker-Walz, Erdmann, Fl{\"o}tter{\"o}d, Hilbrich, L{\"u}cken, Rummel, Wagner, and Wiessner]{lopez2018sumo}
Pablo~Alvarez Lopez, Michael Behrisch, Laura Bieker-Walz, Jakob Erdmann, Yun-Pang Fl{\"o}tter{\"o}d, Robert Hilbrich, Leonhard L{\"u}cken, Johannes Rummel, Peter Wagner, and Evamarie Wiessner.
\newblock Microscopic traffic simulation using {SUMO}.
\newblock In \emph{2018 21st International Conference on Intelligent Transportation Systems (ITSC)}, pp.\  2575--2582, 2018.
\newblock \doi{10.1109/ITSC.2018.8569938}.

\bibitem[Lou et~al.(2022)Lou, Wu, and Ran]{lou2022metarl}
Yican Lou, Jia Wu, and Yunchuan Ran.
\newblock Meta-reinforcement learning for multiple traffic signals control.
\newblock In \emph{Proceedings of the 31st ACM International Conference on Information \& Knowledge Management}, pp.\  4264--4268, 2022.
\newblock \doi{10.1145/3511808.3557640}.

\bibitem[Mei et~al.(2023)Mei, Li, Shi, and Wei]{mei2023missing}
Hao Mei, Junxian Li, Bin Shi, and Hua Wei.
\newblock Reinforcement learning approaches for traffic signal control under missing data.
\newblock In \emph{Proceedings of the Thirty-Second International Joint Conference on Artificial Intelligence}, pp.\  2261--2269, 2023.
\newblock \doi{10.24963/ijcai.2023/251}.

\bibitem[Mercader et~al.(2020)Mercader, Uwayid, and Haddad]{mercader2020maxpressure}
Pedro Mercader, Wasim Uwayid, and Jack Haddad.
\newblock Max-pressure traffic controller based on travel times: An experimental analysis.
\newblock \emph{Transportation Research Part C: Emerging Technologies}, 110:\penalty0 275--290, 2020.
\newblock \doi{10.1016/j.trc.2019.10.002}.

\bibitem[Oroojlooy et~al.(2020)Oroojlooy, Nazari, Hajinezhad, and Silva]{oroojlooy2020attendlight}
Afshin Oroojlooy, Mohammadreza Nazari, Davood Hajinezhad, and Jorge Silva.
\newblock {AttendLight}: Universal attention-based reinforcement learning model for traffic signal control.
\newblock In \emph{Advances in Neural Information Processing Systems}, volume~33, pp.\  4079--4090, 2020.

\bibitem[Pang et~al.(2026)Pang, Wang, Pun, Chen, and Xiong]{pang2026illmtsc}
Aoyu Pang, Maonan Wang, Man-On Pun, Chung~Shue Chen, and Xi~Xiong.
\newblock {iLLM-TSC}: Integration reinforcement learning and large language model for traffic signal control policy improvement.
\newblock \emph{IEEE Transactions on Vehicular Technology}, 75\penalty0 (8):\penalty0 15762--15776, 2026.
\newblock \doi{10.1109/TVT.2026.3674284}.

\bibitem[Pertsch et~al.(2025)Pertsch, Stachowicz, Ichter, Driess, Nair, Vuong, Mees, Finn, and Levine]{pertsch2025fast}
Karl Pertsch, Kyle Stachowicz, Brian Ichter, Danny Driess, Suraj Nair, Quan Vuong, Oier Mees, Chelsea Finn, and Sergey Levine.
\newblock {FAST}: Efficient action tokenization for vision-language-action models.
\newblock In \emph{Robotics: Science and Systems XXI}, 2025.
\newblock \doi{10.15607/rss.2025.xxi.012}.

\bibitem[Qadri et~al.(2020)Qadri, G{\"o}k{\c{c}}e, and {\"O}ner]{qadri2020review}
Syed Shah Sultan~Mohiuddin Qadri, Mahmut~Ali G{\"o}k{\c{c}}e, and Erdin{\c{c}} {\"O}ner.
\newblock State-of-art review of traffic signal control methods: Challenges and opportunities.
\newblock \emph{European Transport Research Review}, 12\penalty0 (1):\penalty0 55, 2020.
\newblock \doi{10.1186/s12544-020-00439-1}.

\bibitem[Ruan et~al.(2024)Ruan, Li, Wei, Jiang, Lu, Xiong, Mao, and Zhao]{ruan2024coslight}
Jingqing Ruan, Ziyue Li, Hua Wei, Haoyuan Jiang, Jiaming Lu, Xuantang Xiong, Hangyu Mao, and Rui Zhao.
\newblock {CoSLight}: Co-optimizing collaborator selection and decision-making to enhance traffic signal control.
\newblock In \emph{Proceedings of the 30th ACM SIGKDD Conference on Knowledge Discovery and Data Mining}, pp.\  2500--2511, 2024.
\newblock \doi{10.1145/3637528.3671998}.

\bibitem[Tang et~al.(2024)Tang, Dai, and Lv]{tang2024arterial}
Yiqing Tang, Xingyuan Dai, and Yisheng Lv.
\newblock Large language model-assisted arterial traffic signal control.
\newblock \emph{IEEE Journal of Radio Frequency Identification}, 8:\penalty0 322--326, 2024.
\newblock \doi{10.1109/JRFID.2024.3384289}.

\bibitem[Varaiya(2013)]{varaiya2013maxpressure}
Pravin Varaiya.
\newblock Max pressure control of a network of signalized intersections.
\newblock \emph{Transportation Research Part C: Emerging Technologies}, 36:\penalty0 177--195, 2013.
\newblock \doi{10.1016/j.trc.2013.08.014}.

\bibitem[Wang et~al.(2021)Wang, Zhang, Di, and Tian]{wang2021roadside}
Lefei Wang, Zhaoyu Zhang, Xin Di, and Jun Tian.
\newblock A roadside camera-radar sensing fusion system for intelligent transportation.
\newblock In \emph{2020 17th European Radar Conference (EuRAD)}, pp.\  282--285. IEEE, 2021.
\newblock \doi{10.1109/EURAD48048.2021.00079}.

\bibitem[Wang et~al.(2024)Wang, Pang, Kan, Pun, Chen, and Huang]{wang2024llmassisted}
Maonan Wang, Aoyu Pang, Yuheng Kan, Man-On Pun, Chung~Shue Chen, and Bo~Huang.
\newblock {LLM-Assisted Light}: Leveraging large language model capabilities for human-mimetic traffic signal control in complex urban environments.
\newblock \emph{arXiv preprint arXiv:2403.08337}, 2024.

\bibitem[Wang et~al.(2025{\natexlab{a}})Wang, Chen, Cai, Pang, Xie, Ma, Xu, Jiang, Wang, Roullet, Chen, Cui, Kan, Lepech, and Pun]{wang2025transsimhub}
Maonan Wang, Yirong Chen, Yuxin Cai, Aoyu Pang, Yuejiao Xie, Zian Ma, Chengcheng Xu, Kemou Jiang, Ding Wang, Laurent Roullet, Chung~Shue Chen, Zhiyong Cui, Yuheng Kan, Michael Lepech, and Man-On Pun.
\newblock {TranSimHub}: A unified air--ground simulation platform for multi-modal perception and decision-making.
\newblock \emph{arXiv preprint arXiv:2510.15365}, 2025{\natexlab{a}}.

\bibitem[Wang et~al.(2025{\natexlab{b}})Wang, Chen, Pang, Cai, Chen, Kan, and Pun]{wang2025vlmlight}
Maonan Wang, Yirong Chen, Aoyu Pang, Yuxin Cai, Chung~Shue Chen, Yuheng Kan, and Man~On Pun.
\newblock {VLMLight}: Safety-critical traffic signal control via vision-language meta-control and dual-branch reasoning architecture.
\newblock In \emph{Advances in Neural Information Processing Systems}, volume~38, pp.\  44266--44297, 2025{\natexlab{b}}.
\newblock \doi{10.52202/085713-1320}.

\bibitem[Wei et~al.(2018)Wei, Zheng, Yao, and Li]{wei2018intellilight}
Hua Wei, Guanjie Zheng, Huaxiu Yao, and Zhenhui Li.
\newblock {IntelliLight}: A reinforcement learning approach for intelligent traffic light control.
\newblock In \emph{Proceedings of the 24th ACM SIGKDD International Conference on Knowledge Discovery \& Data Mining}, pp.\  2496--2505, 2018.
\newblock \doi{10.1145/3219819.3220096}.

\bibitem[Wei et~al.(2019{\natexlab{a}})Wei, Chen, Zheng, Wu, Gayah, Xu, and Li]{wei2019presslight}
Hua Wei, Chacha Chen, Guanjie Zheng, Kan Wu, Vikash Gayah, Kai Xu, and Zhenhui Li.
\newblock {PressLight}: Learning max pressure control to coordinate traffic signals in arterial network.
\newblock In \emph{Proceedings of the 25th ACM SIGKDD International Conference on Knowledge Discovery and Data Mining}, pp.\  1290--1298, 2019{\natexlab{a}}.
\newblock \doi{10.1145/3292500.3330949}.

\bibitem[Wei et~al.(2019{\natexlab{b}})Wei, Xu, Zhang, Zheng, Zang, Chen, Zhang, Zhu, Xu, and Li]{wei2019colight}
Hua Wei, Nan Xu, Huichu Zhang, Guanjie Zheng, Xinshi Zang, Chacha Chen, Weinan Zhang, Yanmin Zhu, Kai Xu, and Zhenhui Li.
\newblock {CoLight}: Learning network-level cooperation for traffic signal control.
\newblock In \emph{Proceedings of the 28th ACM International Conference on Information and Knowledge Management}, pp.\  1913--1922, 2019{\natexlab{b}}.
\newblock \doi{10.1145/3357384.3357902}.

\bibitem[Wei et~al.(2019{\natexlab{c}})Wei, Zheng, Gayah, and Li]{wei2019survey}
Hua Wei, Guanjie Zheng, Vikash~V. Gayah, and Zhenhui Li.
\newblock A survey on traffic signal control methods.
\newblock \emph{arXiv preprint arXiv:1904.08117}, 2019{\natexlab{c}}.

\bibitem[Wei et~al.(2021)Wei, Zheng, Gayah, and Li]{weietal2021survey}
Hua Wei, Guanjie Zheng, Vikash Gayah, and Zhenhui Li.
\newblock Recent advances in reinforcement learning for traffic signal control.
\newblock \emph{ACM SIGKDD Explorations Newsletter}, 22\penalty0 (2):\penalty0 12--18, 2021.
\newblock \doi{10.1145/3447556.3447565}.

\bibitem[Wen et~al.(2025)Wen, Zhu, Li, Zhu, Tang, Wu, Xu, Liu, Cheng, Shen, Peng, Feng, and Tang]{wen2025tinyvla}
Junjie Wen, Yichen Zhu, Jinming Li, Minjie Zhu, Zhibin Tang, Kun Wu, Zhiyuan Xu, Ning Liu, Ran Cheng, Chaomin Shen, Yaxin Peng, Feifei Feng, and Jian Tang.
\newblock {TinyVLA}: Toward fast, data-efficient vision-language-action models for robotic manipulation.
\newblock \emph{IEEE Robotics and Automation Letters}, 10\penalty0 (4):\penalty0 3988--3995, 2025.
\newblock \doi{10.1109/LRA.2025.3544909}.

\bibitem[Wu et~al.(2021{\natexlab{a}})Wu, Wang, Wu, and Wu]{wu2021dynstgat}
Libing Wu, Min Wang, Dan Wu, and Jia Wu.
\newblock {DynSTGAT}: Dynamic spatial-temporal graph attention network for traffic signal control.
\newblock In \emph{Proceedings of the 30th ACM International Conference on Information \& Knowledge Management}, pp.\  2150--2159, 2021{\natexlab{a}}.
\newblock \doi{10.1145/3459637.3482254}.

\bibitem[Wu et~al.(2021{\natexlab{b}})Wu, Zhang, Shen, L{\"u}, Du, and Wu]{wu2021efficientpressure}
Qiang Wu, Liang Zhang, Jun Shen, Linyuan L{\"u}, Bo~Du, and Jianqing Wu.
\newblock Efficient pressure: Improving efficiency for signalized intersections.
\newblock \emph{arXiv preprint arXiv:2112.02336}, 2021{\natexlab{b}}.

\bibitem[Wu et~al.(2023)Wu, Li, Shen, L{\"u}, Du, and Zhang]{wu2023transformerlight}
Qiang Wu, Mingyuan Li, Jun Shen, Linyuan L{\"u}, Bo~Du, and Ke~Zhang.
\newblock {TransformerLight}: A novel sequence modeling based traffic signaling mechanism via gated transformer.
\newblock In \emph{Proceedings of the 29th ACM SIGKDD Conference on Knowledge Discovery and Data Mining}, pp.\  2639--2647, 2023.
\newblock \doi{10.1145/3580305.3599530}.

\bibitem[Xu et~al.(2025)Xu, Wang, Xia, Zhu, Huang, and Xu]{xu2025vlacache}
Siyu Xu, Yunke Wang, Chenghao Xia, Dihao Zhu, Tao Huang, and Chang Xu.
\newblock {VLA-Cache}: Efficient vision-language-action manipulation via adaptive token caching.
\newblock In \emph{Advances in Neural Information Processing Systems}, volume~38, pp.\  182377--182402, 2025.
\newblock \doi{10.52202/085713-5484}.

\bibitem[Yang et~al.(2025)Yang, Wang, Wen, Luo, Zou, Zhang, Wen, and Zhang]{yang2025efficientvla}
Yantai Yang, Yuhao Wang, Zichen Wen, Zhongwei Luo, Chang Zou, Zhipeng Zhang, Chuan Wen, and Linfeng Zhang.
\newblock {EfficientVLA}: Training-free acceleration and compression for vision-language-action models.
\newblock In \emph{Advances in Neural Information Processing Systems}, volume~38, pp.\  45758--45781, 2025.
\newblock \doi{10.52202/085713-1365}.

\bibitem[Yu et~al.(2020)Yu, Liang, Wei, Jin, Huang, Cai, He, and Hua]{yu2020macar}
Zhengxu Yu, Shuxian Liang, Long Wei, Zhongming Jin, Jianqiang Huang, Deng Cai, Xiaofei He, and Xian-Sheng Hua.
\newblock {MaCAR}: Urban traffic light control via active multi-agent communication and action rectification.
\newblock In \emph{Proceedings of the Twenty-Ninth International Joint Conference on Artificial Intelligence}, pp.\  2491--2497, 2020.
\newblock \doi{10.24963/ijcai.2020/345}.

\bibitem[Yuan et~al.(2026)Yuan, Lai, and Liu]{yuan2026collmlight}
Zirui Yuan, Siqi Lai, and Hao Liu.
\newblock {CoLLMLight}: Cooperative large language model agents for network-wide traffic signal control.
\newblock In \emph{The Fourteenth International Conference on Learning Representations}, 2026.
\newblock URL \url{https://openreview.net/forum?id=KeJqoEVOeY}.

\bibitem[Zawalski et~al.(2024)Zawalski, Chen, Pertsch, Mees, Finn, and Levine]{zawalski2024ecot}
Micha{\l} Zawalski, William Chen, Karl Pertsch, Oier Mees, Chelsea Finn, and Sergey Levine.
\newblock Robotic control via embodied chain-of-thought reasoning.
\newblock \emph{arXiv preprint arXiv:2407.08693}, 2024.

\bibitem[Zeng et~al.(2025)Zeng, Yu, Yang, Ao, Hao, Yuan, Li, Wang, and Yang]{zeng2025citylight}
Jinwei Zeng, Chao Yu, Xinyi Yang, Wenxuan Ao, Qianyue Hao, Jian Yuan, Yong Li, Yu~Wang, and Huazhong Yang.
\newblock {CityLight}: A neighborhood-inclusive universal model for coordinated city-scale traffic signal control.
\newblock In \emph{Proceedings of the 34th ACM International Conference on Information and Knowledge Management}, pp.\  4036--4044, 2025.
\newblock \doi{10.1145/3746252.3761285}.

\bibitem[Zhang et~al.(2022)Zhang, Wu, Shen, L{\"u}, Du, and Wu]{zhang2022expression}
Liang Zhang, Qiang Wu, Jun Shen, Linyuan L{\"u}, Bo~Du, and Jianqing Wu.
\newblock Expression might be enough: Representing pressure and demand for reinforcement learning based traffic signal control.
\newblock In \emph{Proceedings of the 39th International Conference on Machine Learning}, volume 162 of \emph{Proceedings of Machine Learning Research}, pp.\  26645--26654. PMLR, 2022.
\newblock URL \url{https://proceedings.mlr.press/v162/zhang22ah.html}.

\bibitem[Zheng et~al.(2019)Zheng, Xiong, Zang, Feng, Wei, Zhang, Li, Xu, and Li]{zheng2019frap}
Guanjie Zheng, Yuanhao Xiong, Xinshi Zang, Jie Feng, Hua Wei, Huichu Zhang, Yong Li, Kai Xu, and Zhenhui Li.
\newblock Learning phase competition for traffic signal control.
\newblock In \emph{Proceedings of the 28th ACM International Conference on Information and Knowledge Management}, pp.\  1963--1972, 2019.
\newblock \doi{10.1145/3357384.3357900}.

\bibitem[Zhou et~al.(2024)Zhou, Liu, Yurtsever, Zagar, Zimmer, Cao, and Knoll]{zhou2023vlmsurvey}
Xingcheng Zhou, Mingyu Liu, Ekim Yurtsever, Bare~Luka Zagar, Walter Zimmer, Hu~Cao, and Alois~C. Knoll.
\newblock Vision language models in autonomous driving: A survey and outlook.
\newblock \emph{IEEE Transactions on Intelligent Vehicles}, 2024.
\newblock \doi{10.1109/TIV.2024.3402136}.

\bibitem[Zhou et~al.(2025)Zhou, Cai, Zhao, Zhang, Huang, Zhou, and Ma]{zhou2025autovla}
Zewei Zhou, Tianhui Cai, Seth~Z. Zhao, Yun Zhang, Zhiyu Huang, Bolei Zhou, and Jiaqi Ma.
\newblock {AutoVLA}: A vision-language-action model for end-to-end autonomous driving with adaptive reasoning and reinforcement fine-tuning.
\newblock In \emph{Advances in Neural Information Processing Systems}, volume~38, pp.\  31725--31761, 2025.
\newblock \doi{10.52202/085713-0942}.

\bibitem[Zou et~al.(2026)Zou, Yang, Chen, Hao, Chen, Huang, and Liang]{zou2026trafficr1}
Xingchen Zou, Yuhao Yang, Zheng Chen, Xixuan Hao, Yiqi Chen, Chao Huang, and Yuxuan Liang.
\newblock {Traffic-R1}: Reinforced {LLMs} bring human-like reasoning to traffic signal control systems.
\newblock In \emph{Proceedings of the 64th Annual Meeting of the Association for Computational Linguistics (Volume 1: Long Papers)}, pp.\  21823--21838, 2026.
\newblock \doi{10.18653/v1/2026.acl-long.995}.

\end{thebibliography}
